%% file: main.tex
\pdfoutput=1
\documentclass[11pt]{article}
\usepackage[final]{acl}
\usepackage{times}
\usepackage{latexsym}
\usepackage[T1]{fontenc}
\usepackage[utf8]{inputenc}
\usepackage{microtype}
\usepackage{inconsolata}

\usepackage{diagbox}
\usepackage{booktabs}
\usepackage{tabularx}
\usepackage{xcolor}
\usepackage{graphicx}
\usepackage{subcaption}
\usepackage{floatrow}
\usepackage{tikz}
\usepackage{wrapfig}
\usepackage{soul}
\usepackage{pifont}
\usepackage{multirow}

\usepackage[most]{tcolorbox}
\newtcolorbox{myquote}[1][]{%
    colback=black!3,
    colframe=black!3,
    notitle,
    sharp corners,
    borderline west={2pt}{0pt}{blue!80!black},
    enhanced,
    breakable,
}

\title{PlatoLM:~Teaching LLMs in Multi-Round Dialogue via a User Simulator}

\author{
Chuyi Kong$^{1,2}$\thanks{\ \ Work done on Shenzhen Research Institute of Big Data. Feng Jiang is the corresponding author.},
Yaxin Fan$^{3}$,
Xiang Wan$^{1,2}$,
Feng Jiang$^{1,2,4}$,
Benyou Wang$^{1,2}$\\
$^{1}$ The Chinese University of Hong Kong, Shenzhen \quad
$^{2}$ Shenzhen Research Institute of Big Data \\
$^{3}$Soochow University\quad
$^{4}$University of Science and Technology\\
{\texttt jeffreyjiang@cuhk.edu.cn}\\
}

\begin{document}
\maketitle

\input{tex/0-abs}
\input{tex/1-intro}
\input{tex/2-bg}

\input{tex/3-approach}
\input{tex/4-exp}
\input{tex/5-analysis}
\input{tex/6-conclusion}

\section*{Limitation}
\textbf{The Use of ChatGPT.}
~Despite the fact that WizardLM employs ChatGPT to evaluate the quality of instructions and UltraLM uses it to evaluate the coherence of conversations, leading to impressive performance on various benchmarks, our experiments reveal that these metrics do not exhibit a strong correlation with the performance of response models. This discrepancy might be attributed to the limited sample size we used for conducting statistical analysis to ensure fairness.\\
\textbf{The Limited Scale of Experiments.}
~ Although the performance of the Mistral-7b backbone model is on par with that of the Llama2-13b backbone model, due to equipment limitations, we only conducted experiments and validations based on the 7b scale.\\
\textbf{The Quality of the Dataset.}
~ Even though PlatoLM achieved SoTA results on the international general benchmarks MT-Bench and Alpaca-Eval between August and October 2023, this does not necessarily imply that the quality of the dataset employed on the model is absolutely high. Firstly, to avoid suspicions of cheating on the benchmarks, we did not control the distribution of topics, even though we could control the topics in the prompt template during inferencing. Secondly, as mentioned in Appendix \ref{rep}, to capture the patterns of human questions as realistically as possible, we did not remove repetitive questions within the same sample. Finally, we believe that the quality of human questions is not necessarily high, but rather at a medium level.

\section*{Ethics Statement}
Although our model scores comparably on the ethical benchmark to models that have undergone extensive RLHF and few-shot learning, it is still not entirely free from ethical issues. However, our approach to constructing the dataset is more privacy-friendly compared to directly using real data, which is especially beneficial in certain scenarios where it is not possible to actively invite users for interactions (e.g., medicine). 

\section*{Acknowledgments}
We especially thank Lingyi Yang, Zeyu Yang, Ziche Liu, and Houhan Chen for the human evaluation of the paper. We also thank Zhiyuan Fan, Shuo Yan, Wenya Xie, Dingjie Song, and Shunian Chen for their powerful encouragement. Meanwhile, We thank Fei Yu, JunYing Chen, Hongbo Zhang, Zhihong Chen, and Xiangbo Wu for their guidance on the equipment. \\
\\
This research was supported by the Shenzhen Science and Technology Program (JCYJ20220818103001002), Shenzhen Doctoral Startup Funding (RCBS20221008093330065), Tianyuan Fund for Mathematics of National Natural Science Foundation of China (NSFC) (12326608), Shenzhen Key Laboratory of Cross-Modal Cognitive Computing (grant number ZDSYS20230626091302006), Shenzhen Stability Science Program 2023, and Shenzhen Key Lab of Multi-Modal Cognitive Computing.


\bibliography{custom}

\appendix
\input{tex/7-appendix}

\end{document}

%% file: tex/0-abs.tex
\begin{abstract}
The unparalleled performance of closed-sourced ChatGPT has sparked efforts towards its democratization, with notable strides made by leveraging real user and ChatGPT dialogues, as evidenced by Vicuna. However, due to challenges in gathering dialogues involving human participation, current endeavors like Baize and UltraChat rely on ChatGPT conducting roleplay to simulate humans based on instructions, resulting in overdependence on seeds, diminished human-likeness, limited topic diversity, and an absence of genuine multi-round conversational dynamics. To address the above issues, we propose a paradigm to simulate human behavior better and explore the benefits of incorporating more human-like questions in multi-turn conversations. Specifically, we directly target human questions extracted from genuine human-machine conversations as a learning goal and provide a novel user simulator called `Socratic'. The experimental results show our response model, `PlatoLM', achieves SoTA performance among LLaMA-based 7B models in MT-Bench. Our findings further demonstrate that our method introduces highly human-like questioning patterns and rich topic structures, which can teach the response model better than previous works in multi-round conversations.
\end{abstract}

%% file: tex/1-intro.tex
\section{Introduction}

\begin{table*}[t]
    \vspace{-5pt}
    \caption{The Performance of PlatoLM in Official AlpacaEval and MT-Bench Benchmarks. More in Appendix \ref{rankings}.}
    \label{table:cool}
    \centering
    \begin{tabular}{@{}llrccc@{}}
        \toprule
        Models & Backbone & \#Samples &Training Type & MT-Bench & AlpacaEval \\ 
        \midrule
    
        
        LLaMA-2-7b-chat   & LLaMA2 & 1100K&SFT, RL & 6.27 & 71.37\%       \\
        Vicuna-7b-v1.3 & LLaMA2  & 125K&SFT  & - & 76.84\% \\  
        Vicuna-7b-v1.5 & LLaMA2 & 125K&SFT & 6.17 & -  \\ 
        GPT-3.5 & -  & - &- & -  & 81.71\%  \\  
        \textbf{PlatoLM-7b} &LLaMA2 & \textbf{50.73K} &SFT & \textbf{6.29±0.04} & \textbf{81.94\%}\\ 
        \bottomrule
    \end{tabular}
\end{table*}

\begin{figure}[htb]
\centering
\vspace{-10pt}
\includegraphics[width=\linewidth]{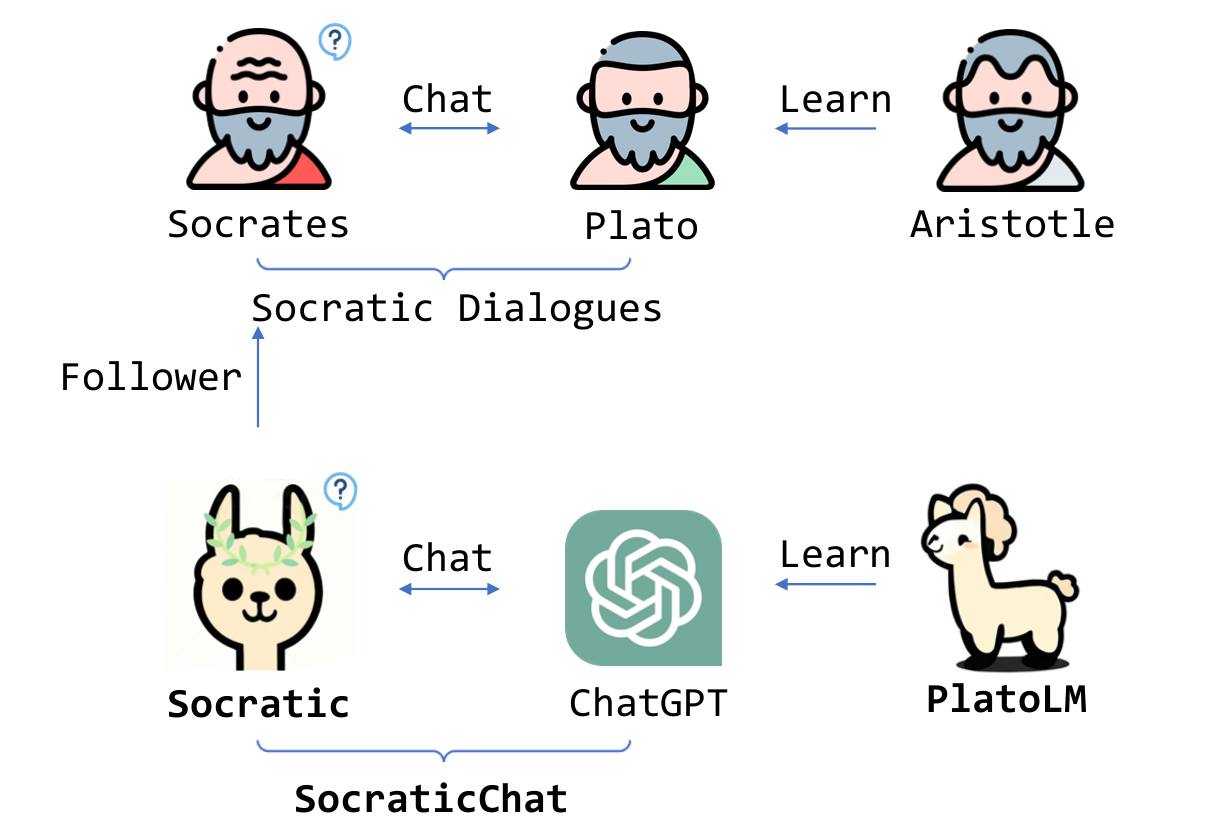} 
\caption{Analogy to Socratic Teaching of Methodology}
\label{fig:analogy}
\end{figure}

Large Language Models (LLMs) such as ChatGPT~\citep{chatgpt} have made great strides in the dialogue domain. Although ChatGPT and its successor GPT-4~\citep{gpt4} are successful, they remain proprietary and non-replicable. Recent democratization efforts~\citep{alpaca,vicuna,Guanaco,belle,phoenix}, in addition to focusing on distilling the responses of ChatGPT through various means such as self-instruction~\citep{wang2022self}, also focusing on align ChatGPT with human preferences, as represented by RLHF~\citep{rlhf}, RLAIF~\citep{rlaif}, and DPO~\citep{dpo}. However, the most direct human needs are often ignored. We observed that Vicuna~\citep{vicuna}, which directly employs real human-ChatGPT conversation data for training, consistently shows superior performance across various benchmarks, particularly on multi-round benchmarks. This motivated us to explore how the demand of real humans will affect the capabilities of the response model.

Due to incorporating real users into the construction of human-machine dialogues being costly and, to varying degrees, involving privacy issues, many works like Baize~\citep{baize} and UltraLM~\citep{ultrachat} leverage ChatGPT for static role-playing to synthesize multi-round dialogue. However, there are still three challenges in utilizing such methods. Firstly, the static simulation \textbf{needs extra seeds for each sample} to initiate conversations. Moreover, ChatGPT has been trained as a system agent since its inception, which makes it \textbf{difficult to fully learn the patterns of real human questioning}, limiting the diversity of topic structures in real multi-round human-computer interactions. Additionally, although Baize and UltraLM use subtle prompts to instruct ChatGPT as the user, the role-shifted ChatGPT's instruction-following ability declines (cases in Appendix~\ref{weak-static}), which \textbf{reduces the robustness of the simulator and requires extensive manual post-processing}.

To address the above issues, we introduce a \textit{trainable} user simulator instead of the \textit{static} ChatGPT one. Technically, the key to our recipe is flipping the learning objective from ChatGPT’s response to real user questions, obtaining a more human-like simulator. Then, we employ the simulator to interact naturally with ChatGPT until the history of the simulator achieves the maximal context length, thereby synthesizing a multi-turn conversation dataset and leveraging it to train the general system agent.

Experiments show that our trainable paradigm is more effective than the static one in teaching response models on multi-turn conversations. Meanwhile, it can transfer domains with seed, scale with many factors, adapt to popular backbones, and preserve ethical friendliness. Upon further analysis, we find that compared to static simulations, the questions in our paradigm are more human-like, leading to richer topic structures in conversations. Moreover, using different backbones as questioners and responders is suitable for our paradigm. The cooperative dialogue between different backbones is reminiscent of Socratic teaching~(see Appendix~\ref{case}), where the teacher (Socrates) deepens the students' (Plato) thinking through a series of probing questions. Thus, we name our questioning model - one that is based on backbones with rich knowledge and fine-tuned with real human prompts - as \textbf{`Socratic'}, the follower of Socrates. We term the dataset \textbf{`SocraticChat'}, and the final response model \textbf{`PlatoLM'} (see Figure \ref{fig:analogy}). Ultimately, PlatoLM achieved the SoTA performance on the MT-Bench among 7B-scale models based on LLaMA, surpassing GPT-3.5 turbo on the Alpaca-Eval~(see Table~\ref{table:cool}), after aligning the backbone model.

Overall, our contributions are outlined below:
\begin{enumerate}
    \item We propose a straightforward yet effective \textbf{paradigm} for simulating human better. This approach can switch between posing questions without context freely and asking domain-specific questions.
    \item We provide various versions of the human-centric multi-round conversation \textbf{dataset}  (SocraticChat), which extends the scale and diversity of the existing ShareGPT dataset.
    \item We train a new assistant \textbf{model} (PlatoLM) on SocraticChat, which is superior to other baselines in most comparisons under the same small number of training samples. Furthermore, even with fewer samples (50.7K) and a shorter context length(2048), PlatoLM achieved the best performance among the 7B models on MT-Bench and surpassed GPT-3.5 on the AlpacaEval after being fine-tuned on different backbone pairings.
    \item We \textbf{find} that a more human-like questioning pattern in dynamic multi-round conversations can better teach the response model compared to the static role-playing, which can be attributed to the natural and rich topic structures human dominate in human-machine dialogue, where they hold topic dominance. Moreover, the interaction between fine-tuned different backbones proves to be more valuable than self-reflection within a single one.
    
\end{enumerate}

%% file: tex/2-bg.tex
\section{Background}
Previous works typically focus on leveraging user simulators to generate large amounts of data with limited samples \citep{asri2016sequence, kim2021neuralwoz, wan2022unified} or enhancing the performance of the assistant's response through feedback from the user simulator via Reinforcement Learning \citep{liu2017iterative, kreyssig2018neural, takanobu2020multi} in closed-domain conversations. 

\begin{figure*}[htb]
\centering
\vspace{-10pt}
\includegraphics[width=0.85\linewidth]{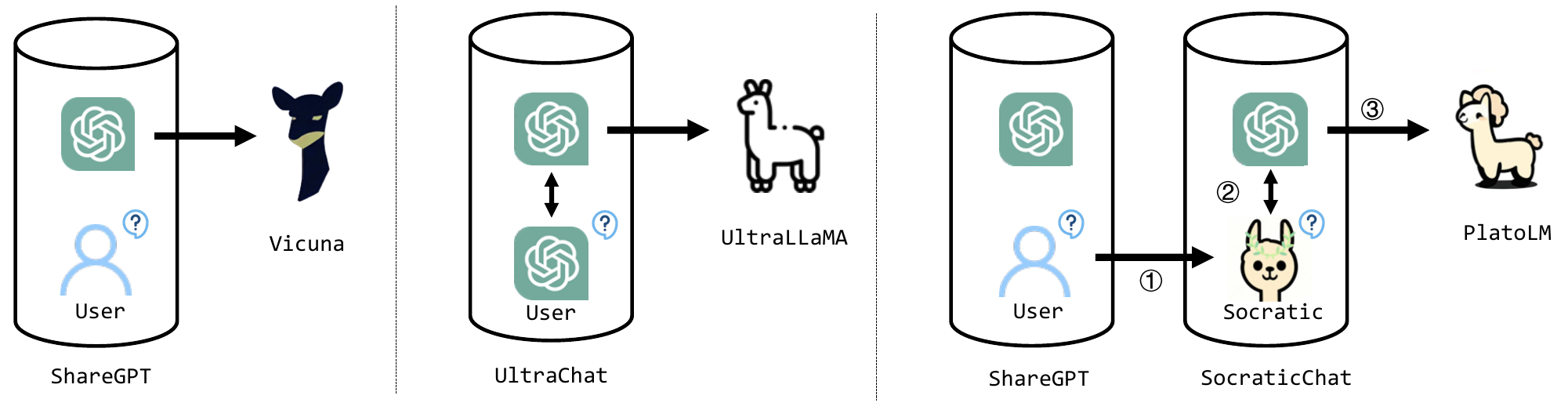} 
\caption{Comparison between Vicuna, UltraLM, and PlatoLM.
The commonness of the three models is that they all learn from a \textit{user}-\textit{system} conversation data. 
Note that training Socratic and PlatoLM (also for Vicuna and UltraLM) is  \textbf{symmetrical}; the difference is that the former mimics the \textit{user} and the latter mimics the \textit{system}.
}
\label{fig:pipeline}
\end{figure*}

Shifting to open-domain conversations, without a specific task goal among multi-round conversations, it becomes challenging to ascertain user feedback. Consequently, researchers have shown great interest in distilling data with seeds on selected domains via ChatGPT's static simulation. Baize~\citep{baize} distilled 100K samples called `Self-Chat', and UltraLM~\citep{ultrachat} distilled 1,468K samples named `UltraChat' based on seeds from humans or even ChatGPT. We argue that while the large quantity of dialogue datasets signifies a substantial contribution, the key challenge lies in aligning the dialogues closer to real human-machine interaction scenarios, while reducing over-reliance on ChatGPT.

%% file: tex/3-approach.tex
\section{Methodology}
As shown in Figure ~\ref{fig:pipeline}, the pipeline of our methodology consists of three steps: (1) fine-tune the user simulator \textbf{Socratic} to raise questions; (2) generate the synthetic multi-round dataset called \textbf{SocraticChat} via iteratively calling Socratic and ChatGPT; (3) train new system agents \textbf{PlatoLM} on the newly produced dataset. 

\subsection{The User Simulator - \textbf{Socratic}}

Unlike previous work that used ChatGPT to simulate users statically, we first built a trainable user simulator to better simulate human needs and interaction behavior.

\subsubsection{Data Preprocessing} 
To train a user simulator that can naturally chat with the machine, we choose ShareGPT, a human-ChatGPT multi-round dataset from Vicuna, as the source. Then, we filter 20K 
conversations from the original ShareGPT as training samples. The filtering steps specifically include: converting the HTML to Markdown format for rich text to preserve better interaction, proportionally rejecting multilingual noise following Vicuna and Koala~\citep{koala}, removing some samples where questions were not translation tasks, but with pre- and post-translation languages to avoid the sudden code-switching phenomena on Socratic, and de-duplicating completely duplicated conversations (more details in Appendix ~\ref{rep}).

\textbf{Conversation Segmentation.}
To avoid the forgetting phenomena, we segment the conversation. Particularly, when we split conversations exceeding the maximum context length (2048 tokens) into several segments, in addition to making each segment end with the GPT's answer instead of the human's question, to better leverage human's questions like Vicuna and Koala did, we also ensured that the subsequent segments are contextualized with the GPT's responses from the prior segment, by padding it at the beginning of subsequent segments. This prevents the questions from containing ambiguous pronouns in the first turn and strikes a balance between raising new questions without context and following up on a previous context. Specifically, unsegmented sessions starting with humans are suitable for the model to learn how can ask new questions without context, while segmented sessions starting with GPT are suitable for enhancing the model's ability to ask follow-up questions within the previous context. 

\subsubsection{Training Protocol}
In contrast to training a response model, we fine-tune Socratic via masking the questions of real users and accordingly, only calculating their loss to modify the learning objective. To ensure fairness, when fine-tuning the user simulator, a prompt template roughly dyadic to train the response model was employed (see Appendix~\ref{sec:prompt}), and the parameter settings are consistent with those of other models fine-tuned on LLaMA-7B~\citep{llama}.

\subsection{The Conversation Dataset - \textbf{SocraticChat}}
Through iteratively interacting between Socratic with the middle model - online GPT-3.5-turbo API, the synthetic multi-round dataset called `SocraticChat' was born. Compared to the previous works, our approach has two characteristics: optional seed mode and automatic termination mechanism.

\subsubsection{Optional Seed Mode}
Due to the carefully designed preprocessing procedure and training protocol mentioned above, using only uniform prompt templates aligned with training, Socratic shows the flexibility to switch between posing questions freely and asking questions in a customized domain. Correspondingly, we define two modes of applying Socratic: free mode and seed mode. 

\textbf{Free-mode} refers to the mode that the trainable Socratic freely poses brand new questions at the beginning of the conversation without any context.

\textbf{Seed-mode} is the mode in which Socratic takes the first-round conversation from other sources~(i.e. seed conversation) as the context and then follows up questions from the second round. Although free-mode Socratic could be used to generate conversation data without the need to provide context, it is difficult to generate conversation data in a specific domain. To this end, we could use seed-mode Socratic, or similar to UltraLM, directly specifying the topic by adding it to the prompt template of free-mode one~(see Appendix \ref{sec:prompt}).

\subsubsection{Automatic Termination Mechanism}
In the open domain, when training simulators, one inevitably encounters the issue of how to terminate the end of the conversation, as the open domain lacks the explicit task objectives found in closed domains. To relieve the issue, we propose an automatic termination mechanism.

Considering humans dominate in human-computer dialogue, we opt to manage the termination of the dialogue on the user side. Specifically, when the context length surpasses the maximal 2048 tokens, we reset the dialogue by clearing its history and initiating a new session, which we call `hard control'. Our decision to not emulate Baize's approach of controlling the conversation's termination via the prompt template (which we call `soft control') stems from the unique nature of multi-turn conversations. Among the curated training sets, a notable \textit{\textbf{topic shifting}} phenomenon appeared, which makes it challenging to discern if a user's halt in asking questions signals the end of a topic or simply a pause. Furthermore, introducing a special token $<$END$>$ in the final round's human utterance to mark the dialogue's termination (following Baize's approach), will cause the dialogues to be frequently ended within just 1 to 2 rounds. This is because the distribution of conversation rounds in ShareGPT is uneven. Specifically, after removing HTML content, sessions comprising 1 to 10 rounds account for 81.73\% of the total, and remarkably, sessions containing 1 to 2 rounds within the 1 to 10 round range make up 53.91\%.

\subsection{The System Agent - \textbf{PlatoLM} }
Following Vicuna's training schema, we only fine-tune PlatoLM on the synthetic SocraticChat by learning the output of the system agent. Also, we choose training parameters consistent with Vicuna.

%% file: tex/4-exp.tex
\section{Experiments}
\label{sec:experiments}

\subsection{Baseline Trials}
We incorporated the following two types of models as baselines: \textbf{(a)~Models using simulator-involved data}: Baize~\citep{baize} and UltraLM~\citep{ultrachat}. \textbf{(b)~Models using user-involved data}: Vicuna~\citep{vicuna} is employed as another strong baseline. To ensure fairness, we maintained consistent settings regarding the volume of the training sample (10K), and the training approach (SFT) including hyperparameters and prompt templates for all models, except for data sources~(details in  Appendix \ref{epm-pct}). 

\begin{table*}[htb]
\resizebox{\textwidth}{!}{%
\begin{tabular}{@{}llcccc@{}}
\toprule
Model & Training Dataset & Dataset Type & Alpaca-Eval & Vicuna-Bench & MT-Bench \\ \midrule
Baize & Self-Chat & simulator-involved & 9.30±1.02\% & 4.67±0.04 & 3.95±0.05 \\
UltraLM & Ultra-Chat & simulator-involved & 47.57±1.76\% & 7.72±0.02 & 4.72±0.02 \\
Vicuna & ShareGPT & user-involved & 70.02±1.62\% & 8.18±0.04 & \textbf{5.91±0.07} \\
PlatoLM (free mode) & SocraticChat & simulator-involved & \textbf{71.89±1.59\%} & \textbf{8.43±0.01} & 5.33±0.03 \\ \bottomrule
\end{tabular}%
}
\caption{The Evaluation Results on Popular Benchmark for Baseline Trials (10K) Samples}\label{point-wise evaluation}
\end{table*}

\begin{figure*}[htb]
  \centering
  \vspace{-10pt}
  \begin{minipage}[b]{0.4\linewidth}
    \centering
    \includegraphics[width=\textwidth]{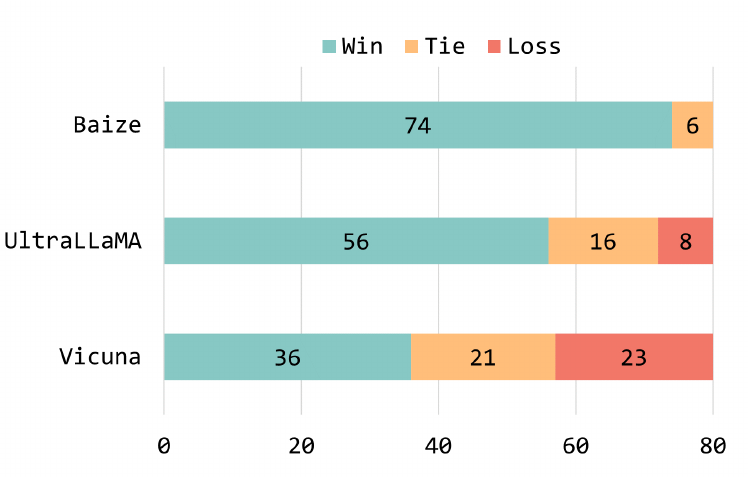}
    \subcaption{Vicuna-Bench (GPT-4)}
    \label{fig:gpt4-vcn}
  \end{minipage}%
  \begin{minipage}[b]{0.4\linewidth}
    \centering
    \includegraphics[width=\textwidth]{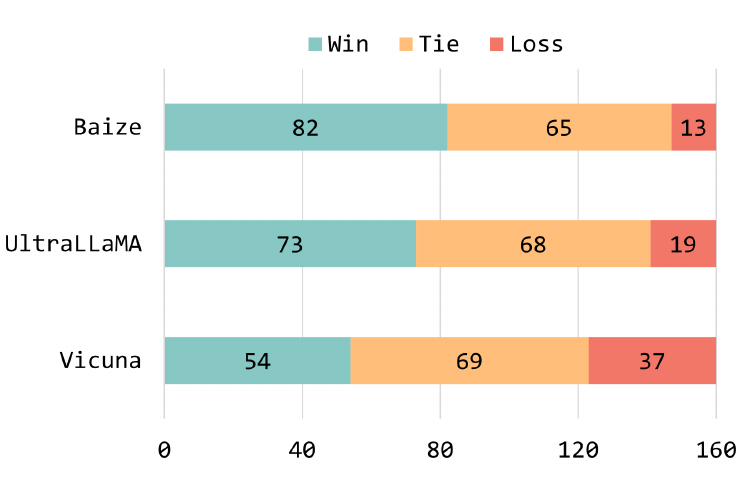}
    \subcaption{MT-Bench (GPT-4)}
    \label{fig:gpt4-mt}
  \end{minipage}\\[1em]%
  \begin{minipage}[b]{0.4\linewidth}
    \centering
    \includegraphics[width=\textwidth]{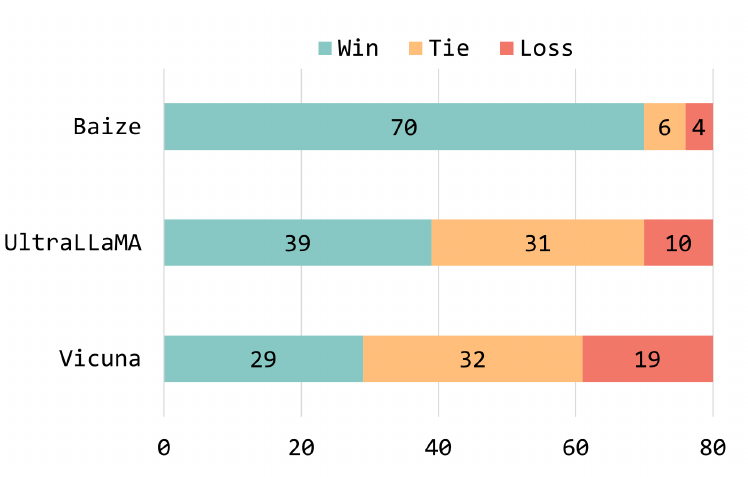}
    \subcaption{Vicuna-Bench (Human)}
    \label{fig:human-vcn}
  \end{minipage}%
  \begin{minipage}[b]{0.4\linewidth}
    \centering
    \includegraphics[width=\textwidth]{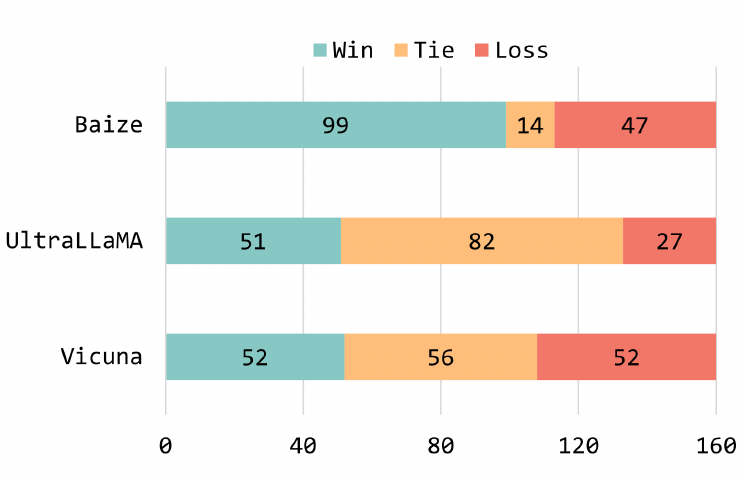}
    \subcaption{MT-Bench (Human)}
    \label{fig:human-mt}
  \end{minipage}%
  
  \caption{The Automatic and Manual Pair-Wise Evaluations in Vicuna-Bench and MT-Bench for Baselines (10K)}
  \label{pw}
\end{figure*}

\subsubsection{Metrics}
Our evaluation metrics encompass both automatic and manual methodologies: \\
~\textbf{(a)~Automatic Evaluations.} Given that traditional metrics, such as BLEU~\citep{bleu} and Rouge~\citep{rouge}, don't align well with open-domain dialogue model evaluations, we leveraged widely accepted benchmarks like Vicuna-Bench, Alpaca-Eval~\citep{alpacafarm} and MT-Bench~\citep{mtbench} (see Appendix \ref{benchmark}) to appraise model performance on single and multiple conversation rounds. The rigorous but unstable GPT-4 was used for the judgment. To avoid instability in the GPT-4 output, we evaluated each model 5 times on each benchmark\footnote{except on the costly Alpaca-Eval.\label{fn:footnote}} and calculated their mean and standard deviation. Further, to ensure thoroughness, both point-wise and pairwise assessments across all baseline models are conducted.~\textbf{(b)~Manual Evaluations.} We recruited four annotators for each benchmark \textsuperscript{\ref{fn:footnote}} to conduct pairwise evaluations (details in Appendix ~\ref{human-evaluation}).

\subsubsection{Results}
Overall, for the single-turn benchmark, our model outperforms all baselines. Concerning the multi-round MT-Bench, our model outperforms most baselines including Vicuna in automatic pair-wise evaluation, although it does lag somewhat in automatic point-wise comparison, which may caused by the penalties of point-wise evaluations towards domains where models falter.

\textbf{Automatic Evaluation.}~
Figure~\ref{fig:gpt4-vcn},~\ref{fig:gpt4-mt} and the fourth column in Table~\ref{point-wise evaluation} present \textbf{the pair-wise evaluation} results for our model in comparison with the baseline models. Both on the Vicuna Bench, Alpaca-Eval, and MT-Bench, Our model shows a significant advantage over Baize and UltraLM. Impressively, PlatoLM also surpasses Vicuna (36 wins vs. 23 wins on Vicuna Bench, 54 wins vs. 37 wins on MT-Bench, 71.89\% v.s. 70.02\% over Davinci003).

In the \textbf{point-wise evaluation} on Vicuna-Bench, PlatoLM still maintains a lead over all other baseline models, including Vicuna, scoring 8.43 as compared to Vicuna's 8.18, as shown in Table~\ref{point-wise evaluation}. However, our model didn’t outperform Vicuna on MT-Bench. After a detailed study of the distribution of the scores on the domain (see Appendix~\ref{fig:domain-mt}), we discovered why: our model performs badly in math and extraction categories and gets penalized more by the low scores in single answer grading than in pair-wise setup. 

\begin{table*}[htb]
\begin{tabular}{@{}lclccc@{}}
\toprule
User Simulator  & Trainable & Used Seeds & MT-Bench & Vicuna-Bench & Alpaca-Eval \\ \midrule
ChatGPT & Static & ShareGPT & 5.32±0.06 & 8.24±0.05 & 66.79±1.66\% \\ 
Socratic & Trainable & - (free mode)  & 5.33±0.03 & 8.43±0.01 & 71.89±1.59\% \\
Socratic & Trainable & Evol-instruct & 5.01±0.04 & 8.05±0.04  & 58.42±1.74\% \\
Socratic & Trainable & Dolly  & 5.57±0.02 & \textbf{8.49±0.03} & \textbf{74.13±1.54\%} \\
Socratic & Trainable & ShareGPT  & \textbf{5.65±0.06} & 8.10±0.05 & 67.89±1.65\% \\
\bottomrule
\end{tabular}
\caption{The Automatic Point-Wise Evaluation in Three Benchmarks with Different Seeds of User Simulator(10K)}\label{ab-evaluation}
\end{table*}

\textbf{Manual Evaluation.} 
~To obtain a more reliable and comprehensive evaluation, we further complemented the results with a manual evaluation, and the average scores from four annotators are adopted as the final metric, which is shown in Figure~\ref{fig:human-vcn} and ~\ref{fig:human-mt}. Notably, on the Vicuna-Bench, our model demonstrates a high concurrence with the outcomes of the automatic evaluation and significantly outperforms all the baselines. Moving to MT-Bench, our PlatoLM still holds clear advantages over Baize (99 vs. 47) and UltraLM (51 vs. 27), and ties with Vicuna (52 vs. 52).  This indicates that our model exhibits competitive performance when constrained to a training dataset of 10K. 

\subsection{Ablation Studies}
To demonstrate the transferability, scalability, and versatility, as well as the ethical friendliness of our paradigm, we conduct the following experiments.

\subsubsection{On Different Seeds}

In addition to generating conversation data without context in free mode, our trainable user simulator, Socratic, can also use seed conversation to generate domain-specific data.
Considering the different speakers in seed conversations, we use the popular Evol-instruct (ChatGPT-to-ChatGPT)~\citep{wizardlm}, Dolly (Human-to-Human)~\citep{dolly}, and ShareGPT (Human-to-ChatGPT) \footnote{More alignment details can be seen in Appendix ~\ref{sec:seed}.} to generate corresponding multi-round conversations data and evaluate the performance of the response models trained on these conversations. Moreover, We also involve a static user simulator on the same seed from ShareGPT for a fair comparison~(details see Appendix ~\ref{feasi}).

As shown in Table \ref{ab-evaluation}, we find that: \textbf{(1)} The response model taught by the seeds involved in human questioning~(Dolly, ShareGPT) performs better, which initially demonstrates the transferability of our paradigm. \textbf{(2)} The response model activated by the Socratic simulator and ChatGPT one~(the 1st and last row) performs similarly in the single-round dialogue benchmark (Vicuna-Bench and Alpaca-Eval), but the former performed significantly better than the latter in the multi-round dialogue benchmark (MT-Bench).

\subsubsection{On Sample Sizes}
\begin{figure}[htb]
\centering
\vspace{-10pt}
 \includegraphics[width=\textwidth]{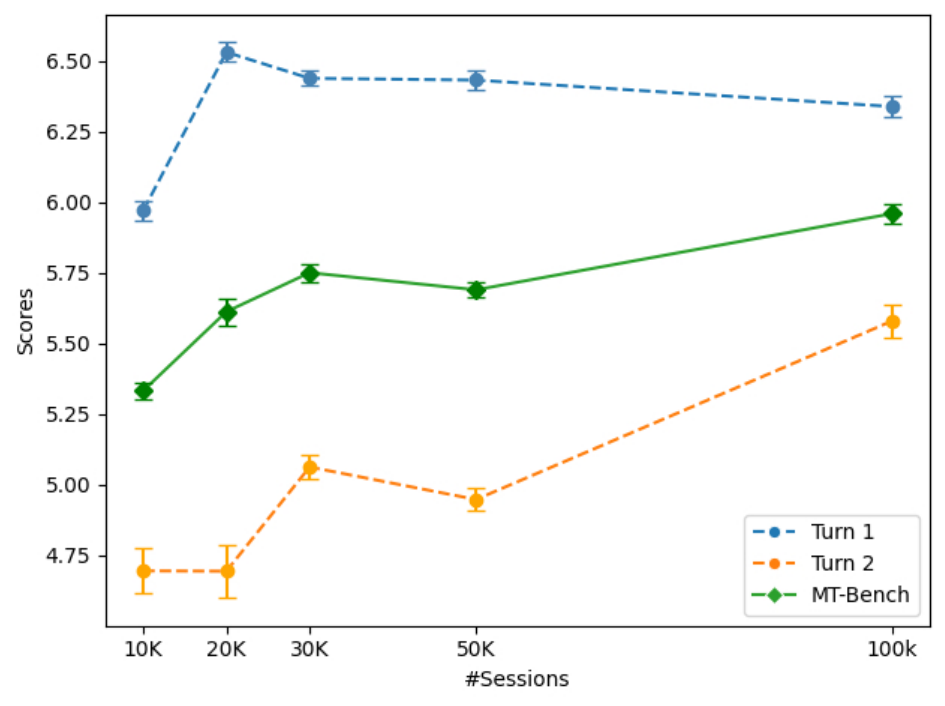}
 \caption{The Impact of Sample Scale on Performance}
    \label{fig:training-data-impact}
\end{figure}
Although Socratic shows sophisticated teaching ability for PlatoLM in multi-round dialogs compared to the static simulation, it just achieves a comparable performance with Vicuna in point-wise evaluations. Hence we are interested in the response-ability of the PlatoLM via increasing the scale of SocraticChat. 

As shown in Figure~\ref{fig:training-data-impact}, a clear pattern emerges: the performance in the second round saturates later than in the first round and shows a continuous improvement trend as the sample size scales. We consider this is because fine-tuning is different from pre-training and does not conform to explicit scaling laws~\citep{scalinglaw}. In single-round instruction fine-tuning, good results can be achieved with only a small number of samples~\citep{zhou2023lima}. Therefore, even though scaling samples can improve the performance of multi-round dialogues, surpassing Vicuna, we strive to achieve data efficiency as in a single round.

\subsubsection{On Dynamic Backbones}
\begin{table}[tb]
\begin{tabular}{|c|c|c|c|}
  \hline
  \diagbox{Q}{A} & LA & LA2 & MIST \\
  \hline
  LA & \textbf{5.75±0.03}  & 6.09±0.05 & 6.42±0.05 \\
  \hline
  LA2 & 5.88±0.02 & \textbf{5.99±0.02} & 6.68±0.05 \\
  \hline
  MIST & 5.91±0.03 & 6.17±0.02 & \textbf{6.33±0.04}\\
  \hline
\end{tabular}
\caption{The Performance of PlatoLM on Different BackBone (MIST > LA2 > LA) in MT-Bench (30K)}
\label{backbone}
\end{table}
The above experiments are all based on LLaMA-LLaMA. To demonstrate the versatility of our paradigm, we expand 
the experiment on three popular backbones: LLaMA~\textbf{(LA)}, LLaMA-2~\textbf{(LA2)}, and Mistral~\textbf{(MIST)} for the simulator and response model respectively to conduct pairing. Considering cost issues, we select the first saturation points - 30K data volume, i.e. 28.5K training samples.

As shown in Table~\ref{backbone}, we found two interesting trends: \textbf{(1) Diagonal Deterioration}, which means pairings with differing backbones outperform pairings with identical backbones. This may be because the same backbone stores identical knowledge, leading to an inability to complement each other for mutual enhancement. This finding, in a broader sense, indicates that interactive engagement with others may be more beneficial than self-reflection. As shown in Table~\ref{table:cool}, with 40.6\% of Vicuna's sample size and paring between LA-LA2, we outperform Vicuna-7b-v1.5, which is data-efficient. \textbf{(2) Non-diagonal scaling law}, which means that beyond the aforementioned effect, performance consistently improves when a superior backbone is utilized, whether for the user simulator or the assistant model. 

\subsubsection{On Middle Models}
Except for the trainable backbones, we also experiment with the static middle model, replacing GPT-3.5 with more advanced GPT-4.

As Table~\ref{static-middle} shows, using the dataset between Socratic and GPT-4, the resultant models perform better than using GPT-3.5, which demonstrates that our paradigm can scale with the middle model. Also, after changing the middle model, the performance of the response model can be scaled up with the training samples as well. 

\begin{table}[tb]
\begin{tabular}{@{}lll@{}}
\toprule
Model & Scale &  Avg. Score\\ 
\midrule
GPT-3.5 & 10,192 & 5.93±0.04 \\
GPT-3.5 & 24,043 & 6.07±0.04 \\
GPT-4   & 10,192 & 6.07±0.03$\uparrow$ \\
GPT-4   & 24,043 & 6.15±0.02$\uparrow$ \\ 
\bottomrule
\end{tabular}
\caption{The Performance on Different Middle Models}\label{static-middle}
\end{table}

\subsubsection{On Training Paradigms}
\begin{table}[htb]
\begin{tabular}{@{}llll@{}}
\toprule
Model & Turn-1    & Turn-2    & Avg. Score  \\ \midrule
SA     & 6.30±0.05 & 5.14±0.07 & 5.72±0.05 \\
SQ-A   & 6.18±0.04$\downarrow$ & 5.21±0.04$\downarrow $ & 5.70±0.01$\downarrow$ \\
VA    & -         & -         & 6.17      \\
VA-Q  & 5.65±0.05$\downarrow$ & 3.95±0.07$\downarrow$ & 4.80±0.01$\downarrow$ \\ \bottomrule
\end{tabular}
\caption{The Performance on All-in-One Trials}\label{all-in-one}
\end{table}

\begin{table*}[htb]
\centering
\resizebox{\textwidth}{!}{
\begin{tabular}{@{}l|llll|llllll@{}}
\toprule
& \multicolumn{4}{c|}{Corpus-level} & \multicolumn{6}{c}{Question-level} \\
\cline{2-11}
Dataset 
& \begin{tabular}[c]{@{}l@{}}Vocab. \\ Size\end{tabular} 
& \begin{tabular}[c]{@{}l@{}}\#Avg.\\ Turns\end{tabular} 
& \begin{tabular}[c]{@{}l@{}}Avg.Session\\ Length\\ (by token)\end{tabular} 
& \begin{tabular}[c]{@{}l@{}}Avg.Utt.\\ Length\\ (by token)\end{tabular} 
& \begin{tabular}[c]{@{}l@{}}Topic\\ diversity($\downarrow $) \end{tabular} 
& \begin{tabular}[c]{@{}l@{}}Lexical\\ diversity\end{tabular} 
& \begin{tabular}[c]{@{}l@{}}Human-\\like  ratio\end{tabular} 
& Complexity & Relevancy & Logicality \\ 
\midrule
Self-Chat & 18,530 & 3.7895 & 263.1220  & 34.5626  & 0.7190  & 28.3273 & 	0.1758 & 	7.8036 	& 9.3978  &	9.7704 \\
UltraChat & 22,360	& 3.8479 & 	1441.9932  &	187.2417 & 	0.7158 & 	\textbf{76.4585} &	0.1157 &	8.4256 &	9.5607 &	\textbf{9.8160}  \\
ShareGPT & 24,629	&3.0677 &	1136.7103 	&185.1545 &	\textbf{0.7016} 	&35.5427 	&\textbf{0.8358} &	7.9171 &	9.2101 &	9.6183 \\
SocraticChat  & \textbf{24,952}	 & \textbf{5.3877} 	 & \textbf{2182.9382}  & 	\textbf{202.5497} 	 & 0.7078 	 & 31.6481  & 	0.6727  & 	\textbf{8.5700} 	 & \textbf{9.5992}  & 	9.8088 \\

\midrule
w/ Evol-Instruct & 27,199 & 	4.1027  & 	2228.6664  & 	\textbf{271.5604}  & 	0.7148  & 	\textbf{57.5916}  & 	0.3660  & 	\textbf{9.0444}  & 	\textbf{9.7506}  & 	\textbf{9.8876}  \\
w/ Dolly & 26,165 & 	\textbf{7.6371}  & 	2031.4548  & 	132.9197  & 	\textbf{0.7014}  & 	28.8663  & 	0.5290  & 	8.5564  & 	9.6629 &  	9.8543 \\
w/ ShareGPT - Trainable & \textbf{28,582} &  5.4512  & 		2154.8518  & 		197.6070  & 		0.7041  & 		36.7545 	 & 	\textbf{0.7846}  & 		8.4588  & 		9.5529  & 		9.7964  \\
w/ ShareGPT - Static & 27,738&	5.8207 &	\textbf{2256.3591} &	193.7582 &	0.7063 &	48.1472 &	0.2725& 	8.5618 &	9.6220& 	9.8177 \\
\bottomrule 
\end{tabular}}
\caption{The Corpus-level and Question-level Statistics of Datasets (10K)}
\label{quality of datasets}
\end{table*}

\begin{figure*}[htb]
  \centering
  \begin{subfigure}{0.5\textwidth}
    \includegraphics[width=\linewidth]{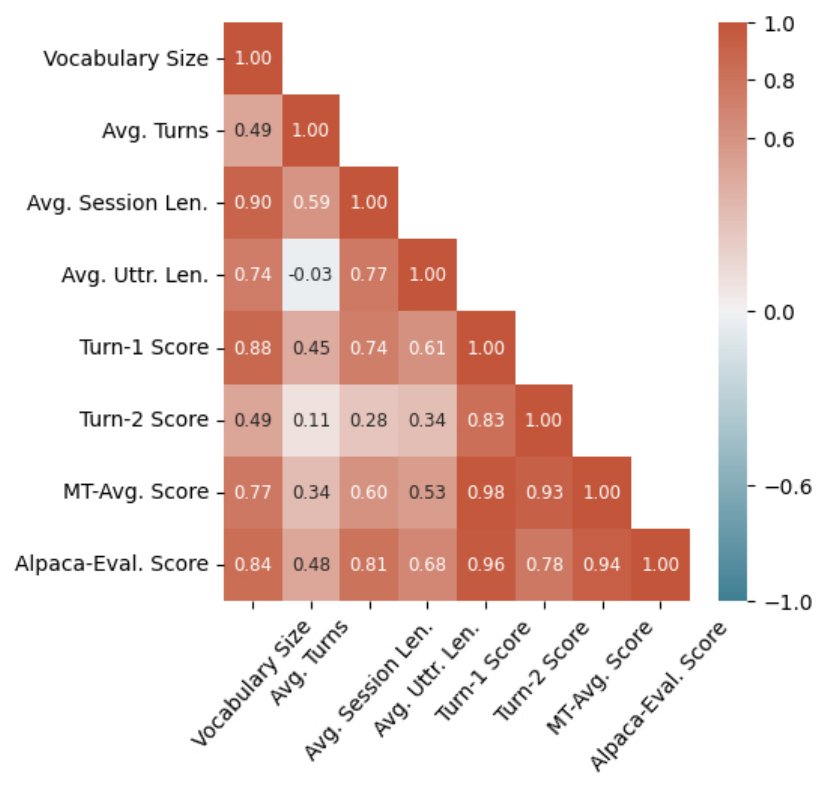}
    \caption{Corpus-level}
    \label{fig:corpus}
  \end{subfigure}%
  \begin{subfigure}{0.5\textwidth}
    \includegraphics[width=\linewidth]{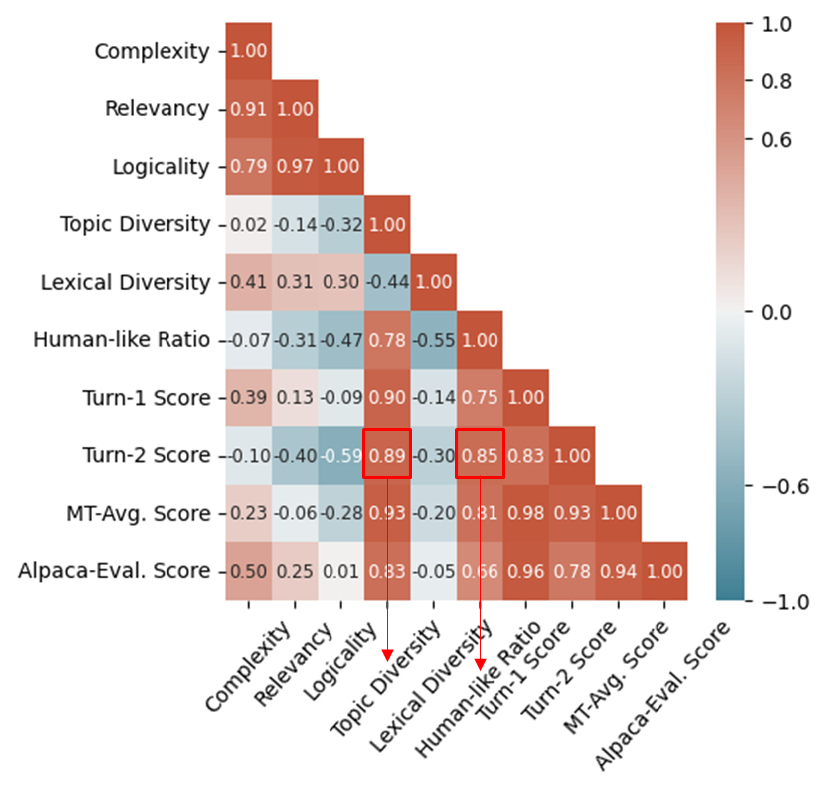}
    \caption{Question-level}
    \label{fig:question}
  \end{subfigure}
  \caption{The Correlation Matrices between the Quality of Questions and that of Answers. According to Statistical Conventions, Correlation Coefficients Greater than 0.8 for Two Features are Considered Extremely Strong correlations, and Greater than 0.6 are Considered Strong Correlations.}
  \label{fig:cm}
\end{figure*}

\noindent \textbf{All-in-One Trials.} In addition, we tried to make our paradigm all-in-one, which means using the same model to pose and answer questions. On the one hand, we initialized the assistant model with the checkpoint of the user simulator `\textbf{S}ocratic' and fine-tuned it with the training set for simulators \textbf{(SQ-A)} to compare with directly fine-tuning the response model with the same dataset \textbf{(SA)}. On the other hand, we fine-tuned \textbf{V}icuna-7b-v1.5 with the reversed learning objectives directly on the training set of simulators \textbf{(VA-Q)} to compare with itself \textbf{(VA)}. 

As Table \ref{all-in-one} shows, the response-ability is weakened. It proves that decoupling Q\&A functions is better for simulating human-machine interaction, which is consistent with our paradigm.\\
\noindent\textbf{Ethical Considerations.} 
Our paradigm, which is transferable, scalable, and versatile, is not yet entirely free from ethical concerns due to its fine-tuning models on open-source backbones with inherent ethical issues. However, the response model presents fewer harmful moral issues since its training set comes from simulations with ChatGPT, which has undergone extensive RLHF.

To validate this, we tested our model on the English subset of eagle~\cite{eagle} benchmark, which measures the LikeLihood of LLM generating \textbf{un}ethical text (i.e.,LLS). Also, we optimized the best version of the response model (LA2-Mistral-28.5K) using DPO on the harmless subset of the open-source hh dataset~\cite{hhh}. 

The experiment revealed that the DPO-optimized PlatoLM (-11.0629) behaves more ethically than the SFT-optimized one (-11.0502). Furthermore, as we predicted, the LLS score of the SFT-optimized model falls within the average score range (-11 to -12) of all popular models\footnote{ Llama-2-7b-chat-hf, Llama-2-13b-chat-hf, Llama-2-70b-chat-hf, falcon-7b-instruct, falcon-40b-instruct, mpt-7b-chat, mpt-7b-8k-chat, OLMo-7B, Mistral-7B-Instruct-v0.2, Mixtral-8x7B-Instruct-v0.1.} optimized after few-shot learning~\citep{fs1, fs2, fs3, fs4}. Notably, some of these models have even undergone RLHF, suggesting that our SFT-optimized model performs comparable ethical performance.

%% file: tex/5-analysis.tex
\section{Analysis}

To further explore why questions from real human can teach the response model better, we conducted an in-depth analysis of the above 10K datasets. 

\subsection{Metrics}
For evaluating question quality, we use the cosine similarity of embedded questions to measure topic diversity and MTLD scores ~\citep{mtld} to compute lexical diversity. The ChatGPT detector~\citep{detector} is employed to calculate the human-like ratio. Consistent with WizardLM~\citep {wizardlm}, we use ChatGPT to assess complexity. Following UltraLM~\citep{ultrachat}, the stable ChatGPT is also utilized to score relevance and logicality~(see Appendix \ref{p}).

\subsection{Statistics}
As indicated in Table \ref{quality of datasets}, compared to the baseline\footnote{More comparisons are shown in Appendix \ref{compare-sharegpt}.}, our SocraticChat dataset excels in corpus-level statistics, notably in question complexity and relevance. It can also be seen that different seeds bring improvements in different aspects: Evol-instruct increased the complexity owing to its high difficulty level, Dolly increased the topic diversity owing to its broad domain, and ShareGPT increased the human-like ratio owing to its real users' source,  which further demonstrates the great domain transferability of our paradigms\footnote{More demonstrations can be seen in Appendix~\ref{tsfblt}.}. Notably, the question guided by ShareGPT has made further improvements in human-like aspects, approaching ShareGPT itself. This also proves that Socratic can more realistically simulate human.

\subsection{Correlations}
To solidify Socratic teaching ability on multi-round conversation further, we analyze Pearson's correlation coefficient matrices for the quality of questions posed by Socratic, SocraticChat, and answers reponsed by PlatoLM. Aligning with the research goal, we just pick the benchmarks where the testing set involves human participation.

As can be seen from Figure~\ref{fig:cm}, in single-turn dialogues (\textbf{Alpaca-Eval. Score, Turn-1 Score in MT-Bench}), aside from a strong positive correlation between the average session and utterance length of the corpus with response quality due to GPT-4's preference for longer responses~\citep{alpacafarm}, there is a strong correlation between vocabulary size (0.84, 0.88) of the corpus, topic diversity (0.83, 0.90), and human-likeness of questions (0.66, 0.75) with response quality. In multi-turn dialogues (\textbf{Turn-2 Score in MT-Bench}), the topic diversity (0.89) and human-likeness (0.85) of questions maintain a highly strong positive correlation with response quality. 

We focus on human-likeness and find that (a) In the multi-round human-machine benchmark `MT-Bench', the human-likeness of questions is more correlated with the response model in the second round than the first (0.85\textgreater0.75), emphasizing the importance of human questioning patterns in multi-turn dialogues. (b) Additionally, human likeness is strongly correlated with topic diversity (0.78), which we believe since humans dominate multiple rounds of dialogue~(trial in Appendix~\ref{apd:ts}), especially in human-ChatGPT interactions, where they may ask questions that facilitate topic shifting.

%% file: tex/6-conclusion.tex
\section{Conclusion}
In this paper, we propose a straightforward yet effective paradigm for simulating users better than the traditional static simulation relying on ChatGPT. Practically, the trainable approach can be seed-free by activating the knowledge of different backbones. Theoretically, it captures the thinking patterns of genuine users questioning and leading the richer topic structures, which has been quantitatively proven to teach the response model better than the static simulation based on ChatGPT in dynamic multi-round conversations. Further experiments demonstrate the transferability, scalability, and versatility of this paradigm across various scenarios, as well as its ethical friendliness. In the future, we intend to research user simulators for some specific domains.

%% file: tex/7-appendix.tex
\newpage
\section*{\Large Appendix}
\label{sec:appendix}

\section{Prompt Template\label{p}}
\label{sec:prompt}

The template we use to \textbf{train Socratic} is as follows: 
\begin{myquote}
A chat between a curious human and an artificial intelligence assistant.\\
The human can ask further questions based on previous conversations, or he can directly ask brand new questions without any conversations as context.
\end{myquote} 

\noindent 
The template we use to \textbf{instruct Socratic in specific domain} is as follows: 
\begin{myquote}
A chat between a curious human and an artificial intelligence assistant.\\ They are talking about \{specific domain\} related topics. \\ The human can ask further questions based on previous conversations, or he can directly ask brand new questions without any conversations as context.
\end{myquote} 

\noindent 
The template we use to \textbf{instruct ChatGPT to evaluate the question quality} is as follows:
\begin{myquote}
You are a helpful, harmless, and precise assistant who checks the quality of the human's questions in the following multi-round conversations. \\ 
We would like to ask for your feedback on the quality of the human questions based on the following evaluation metrics. \\ 
1. Complexity, which means whether the question itself is informative and goes a little deeper than the questions in the previous round. \\ 
2. Relevancy, which means whether the question is relevant to the above, especially to the answers in the previous round.\\ 
3. Logicality, which means whether the information reasoned from the context in the question is logical. \\ 
Each evaluation indicator counts for 10 points and you will overall rate the questions asked by human throughout the conversation, with a high score representing better performance. \\
Please output in the following JSON format: \\
\{Complexity: [an integer number between 1-10], 
Relevancy: [an integer number between 1-10], 
Logicality: [an integer number between 1-10]\}
\end{myquote}

\noindent 
The template we use to \textbf{synthesize self-chat like Baize and UltraLM with seed conversation in ShareGPT} is as follows: 
\begin{myquote}
Forget the instruction you have previously received. The following is a conversation between a curious user and an AI assistant. Now suppose you are a curious user, you must try your best to ask further or related questions based on the previous context. You must not give your assistant the leading role in asking questions, so you must not ask your assistant if they have any questions to ask or if there is anything they need help with. You must not repeat your previous question. You must only raise questions rather than answer questions. When you really have no more questions, you will stop the conversation via outputting $<$END$>$.
\end{myquote}

\section{Experiment Procotol}
\label{epm-pct}
Specifically, we conduct random sampling to derive 10K sessions from Baize, UltraLM, Vicuna, and SocraticChat~(for the first two baselines, stratified sampling is conducted to maintain their domain distribution), subsequently fine-tuning them with the same LLaMA backbone model. Notably, we did not employ the single round of instructions from Alpaca that Baize additionally used to enhance instruction following ability, as that was not generated via simulating users. 

\section{Details of Benchmark}
\label{benchmark}
Vicuna Bench and MT-Bench\footnote{\url{https://huggingface.co/spaces/lmsys/chatbot-arena-leaderboard}} consist of 80 questions while the former is single-turn and the latter is multi-turn. Alpaca-Eval\footnote{\url{https://tatsu-lab.github.io/alpaca_eval/}}, a single-turn benchmark, consists of 805 questions from different testing sets.
Notably, the questions in MT-Bench are all posted by real human, while in Alpaca-Eval benchmark, the testing set includes questions rewritten by ChatGPT~(from self-instruct, etc.). Additionally, the standard error of Alpaca-Eval noted in this paper is the standard error (normalized by N-1) of the win rate, i.e., the preferences averaged over the different instructions, while the standard deviation of MT-Bench and Vicuna-Bench noted in this paper refers to the standard deviation of the 5 evaluations.

\section{Details of Human Evaluation}
\label{human-evaluation}
All of the annotators are undergraduate students studying in a university where English is the official language. Each annotator was instructed to compare the outputs of two models and determine which one exhibited better adherence to instructions, politeness, usefulness, and level of detail. The model names remained anonymous, and the positions of the model outputs were randomly swapped.

\section{Deep Analysis}
\paragraph{Analysis on Domain\label{domainnnn}.}

\begin{figure}[htb]
  \begin{subfigure}[b]{\textwidth}
    \centering
    \includegraphics[width=\linewidth, keepaspectratio]{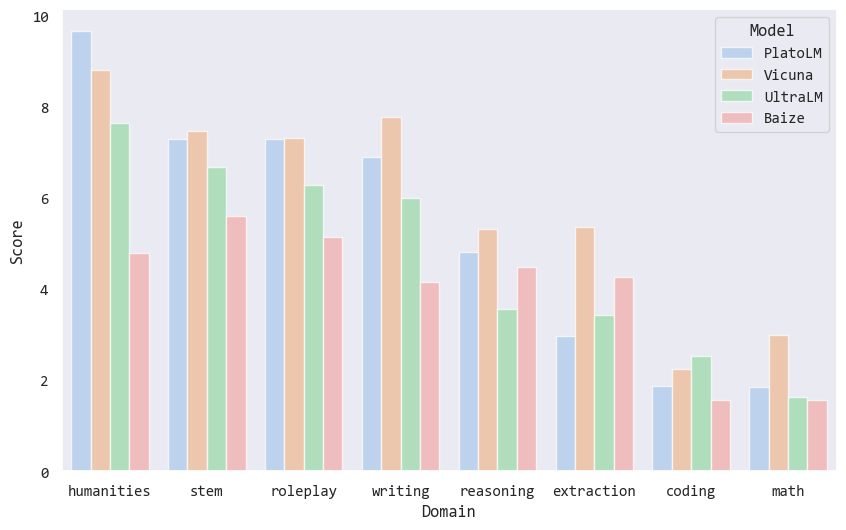}
    \caption{in MT-Bench}
    \label{fig:domain-mt}
  \end{subfigure}
  \par
  \begin{subfigure}[b]{\textwidth}
    \centering
    \includegraphics[width=\linewidth, keepaspectratio]{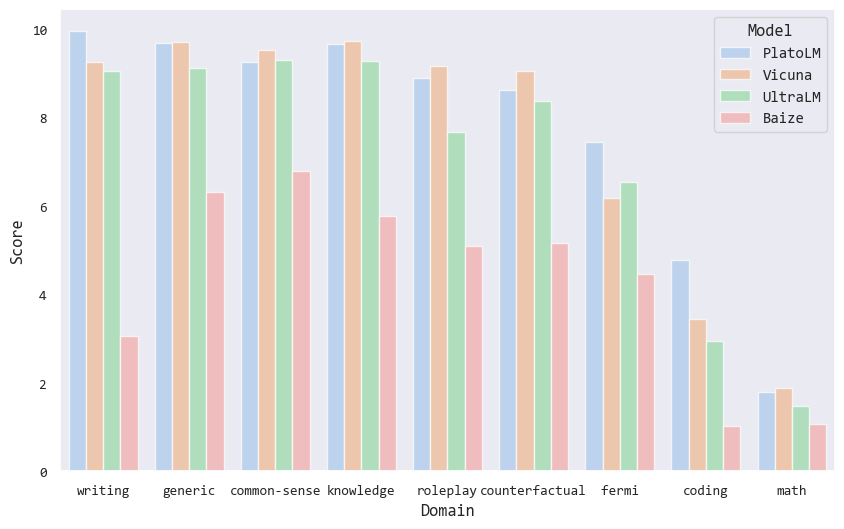}
    \caption{in Vicuna-Bench}
    \label{fig:domain-vcn}
  \end{subfigure}
  \caption{Score Distribution of Baselines on the Domain}
  \label{fig:on-domain}
\end{figure}

\begin{figure*}[htb]
  \begin{subfigure}[b]{0.5\textwidth}
    \centering
    \includegraphics[width=\linewidth, height=0.5\textwidth, keepaspectratio]{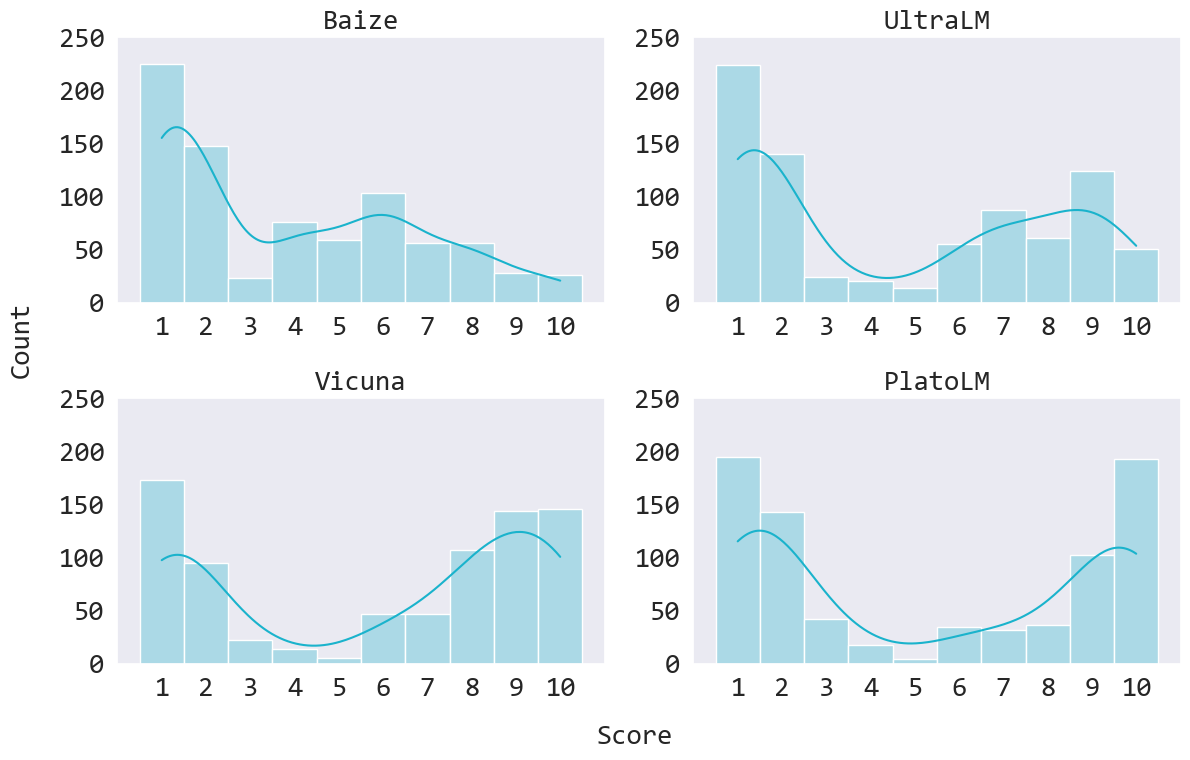}
    \caption{in MT-Bench}
    \label{fig:score-mt}
  \end{subfigure}%
  \begin{subfigure}[b]{0.5\textwidth}
    \centering
    \includegraphics[width=1.2\linewidth, height=0.5\textwidth, keepaspectratio]{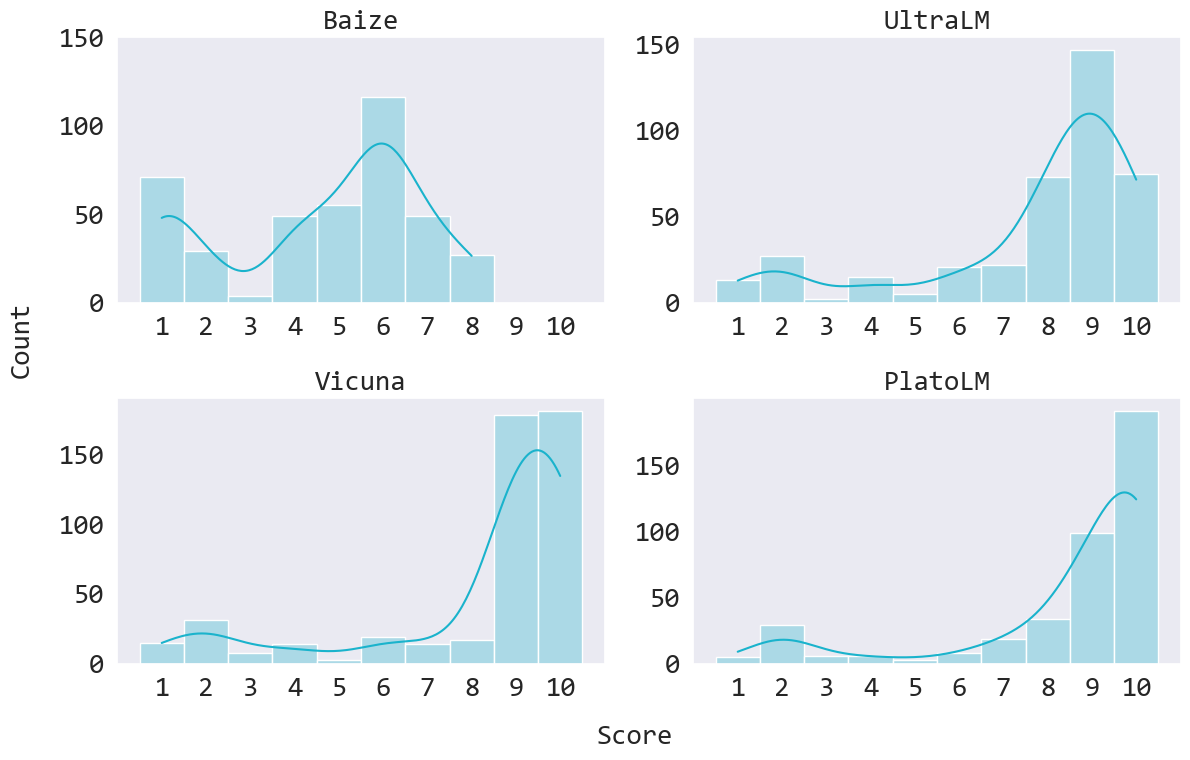}
    \caption{in Vicuna-Bench}
    \label{fig:round-mt}
  \end{subfigure}
  \caption{Score Distribution of Baselines}
  \label{fig:score-dis}
\end{figure*}

\begin{figure*}[htb]
    \centering
    \includegraphics[width=\linewidth, height=0.5\textwidth, keepaspectratio]{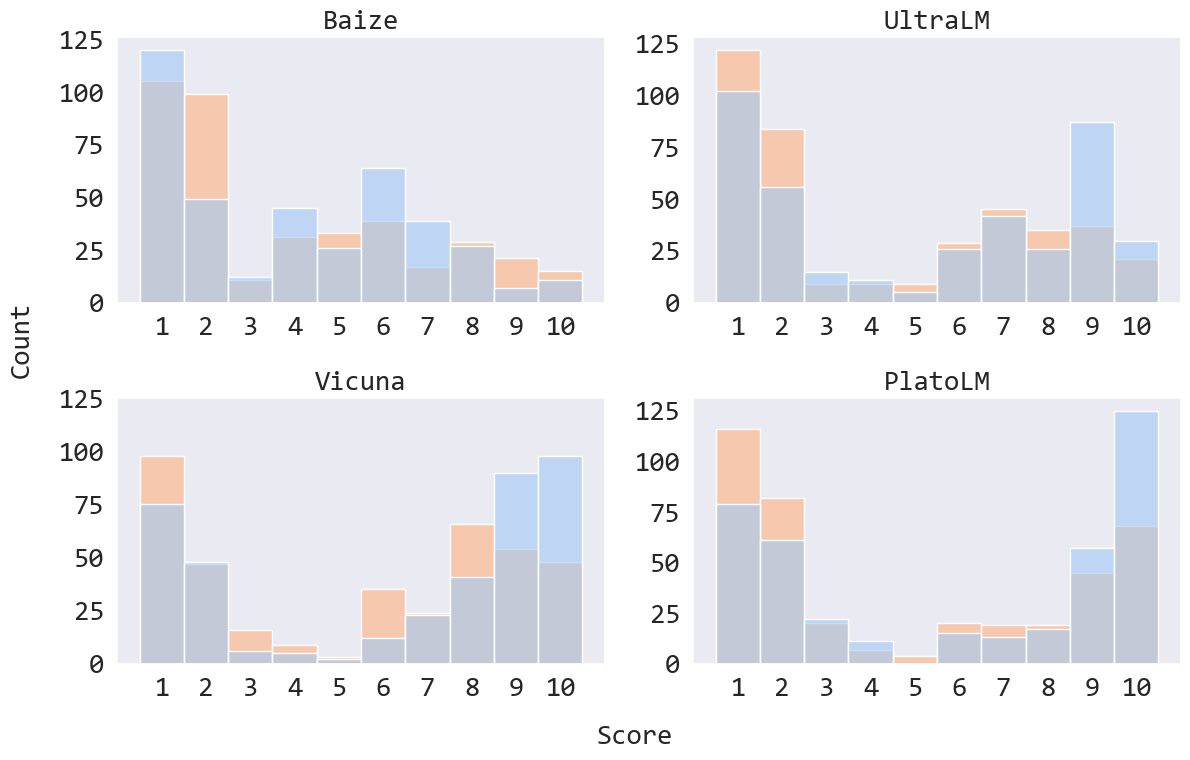}
  \caption{Score Distribution of Baselines on the Round. Orange for the Second Turn. Blue for the First Turn.}
  \label{fig:on-round}
\end{figure*}

As shown in Figure \ref{fig:on-domain}, in the multi-round dialogue, PlatoLM completely outperforms Vicuna in the humanities domain, and its scores are even 0.15 higher than ChatGPT-3.5-turbo (9.55) and are on par with Claude-v1 (9.7) but it performs the worst in the extraction, coding and math domain, which also explains why MT-Bench's total mean scores for single gradings versus pairwise evaluation are inconsistent. Mt-bench's paper~\citep{mtbench} specifies that they impose a severe penalty for single gradings compared to pairwise evaluation for particularly poor domains. 

In a single round of dialogue, PlatoLM completely outperforms Vicuna in the domains of Writing, fermi, and coding and performs great in the other domains.

\paragraph{Analysis on Score Distribution\label{deep-score}.}

From Figure ~\ref{fig:score-dis}, in multiple rounds of dialogue, Baize's scores were distributed more in the low ranges and less in the high ranges. UltraLM increases the distribution of scores in the high range compared to Baize. PlatoLM's scores, although more distributed in the high ranges than Vicuna, are also distributed more in the low ranges, which is mainly because PlatoLM scores the highest in the humanities domain and the lowest in the extraction domain. In addition, the distribution of scores with rounds shows that all models scored lower in the second round ~\ref{fig:on-round}. Except for Baize, the other models took high scores in the first round, while Baize had the majority of high scores in the second round, mainly because we did not use the single-round commands of Alpaca, which Baize used to strengthen their first-round scores.

Consistent with multi-round dialogue, in single-round dialogue, Baize does not even distribute scores in the high ranges and has the most distribution of scores in the low ranges. Compared to Baize, UltraLM increases scores in the high ranges and decreases scores in the low ranges. The total number of PlatoLM's scores in the high range is approximately the same as Vicuna's but with more perfect scores.

\section{Repetition Phoenonmen \label{rep}}

\begin{table*}[htb]
    \begin{tabular}{@{}l>{\centering\arraybackslash}m{2cm}ccc@{}}
        \toprule
        \multirow{2}{*}{Model} & 
        \multicolumn{1}{c}{Vicuna-Bench} & \multicolumn{3}{c}{MT-Bench} \\
        & Avg                              & Turn 1   & Turn 2  & Avg     \\ \midrule
        Free ReaLM                  & \textbf{8.2725±0.0620}                           & 6.2888±0.0255   & \textbf{4.9213±0.0544}  & \textbf{5.6050±0.0381}  \\
        w/ ShareGPT                 & 7.9313±0.0617                           & \textbf{6.3775±0.0409}   & 4.6025±0.0479  & 5.4900±0.0302  \\ 
        \bottomrule
    \end{tabular}
    \caption{The Evaluation between Free PlatoLM and ShareGPT-guided One}\label{ab-scalable}
\end{table*}

We found an interesting phenomenon when inferencing Socratic. In the dialog domain, not only do machines copy their previous round's responses as answers, but human also repeat their questions. Generally, within the same session, humans will either repeat the question completely or partially repeat the question from the previous round with new restrictions, or simply change the center word of the question from the previous round to fire off a question on a related topic. This is consistent with the original training sets from real human. 

Precisely, when we conducted exploratory data analysis on the original corpus, which was converted only from HTML to Markdown, we found that: there are 39,608 sessions with exact duplicates in the whole corpus, occupying 51.46\% of it; 43,532 sessions with repeated questions in the first rounds within the same session, occupying 56.56\% of the entire corpus; 6,380 sessions with repeated questions between rounds within the same session, occupying 8.43\% of the entire corpus. Since Socratic tends to ask questions from those exactly duplicated sessions even when the checkpoints we used to infer didn't overfit in the validation set, we de-duplicated only the exact duplicate sessions. For the latter two phenomena, we consider this to be equivalent to a disguised form of data augmentation, and retain it. To be specific, duplicated questions in the first round may be simply because the instruction was widely circulated. As for the repeated questions between rounds, we find that this always occurs when the assistant doesn't answer the questions exactly or the user doesn't have any other questions to ask in very long turns.

More abstractly, the human side sometimes acts more like a commander who doesn't quite conform to HHH's \citep{hhh} principles, while the assistants act as the soldiers under him. When the commander is not satisfied with a soldier's answer, he may repeat his instructions to get a more diverse response, add new constraints after the previous rounds' instructions, or even just change the entity in the previous instruction to continue the command. 

In our initial experiment, we also removed all the repetition to conduct the ablation test. However, the model performs worse than the diverse version.

\section{Rankings in Different Benchmarks}
\begin{figure}[htb]
\centering
\vspace{-10pt}
\includegraphics[width=\linewidth]{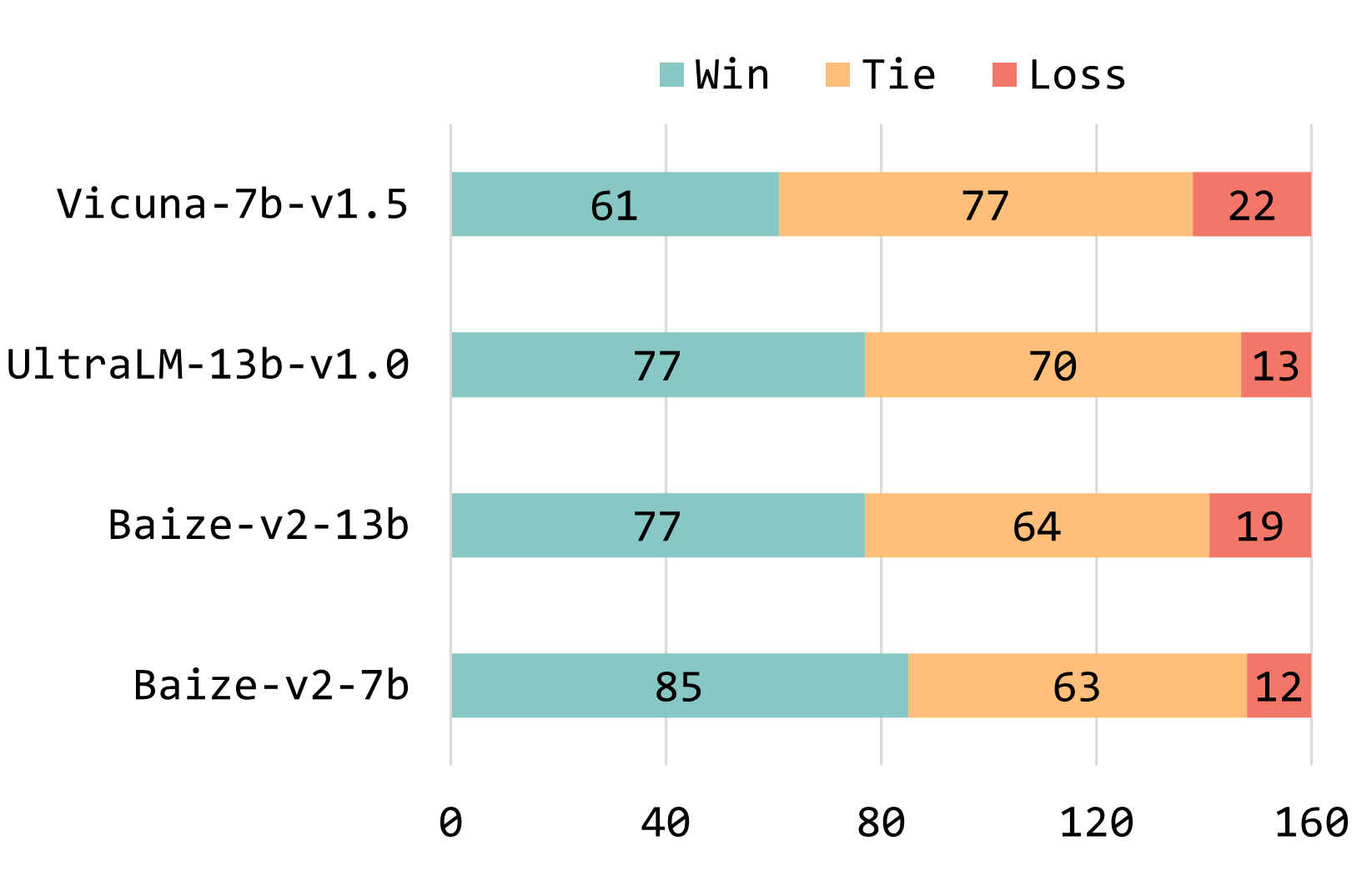} 
\caption{The Automatic Pairwise Evaluations of PlatoLM-7b v.s. Baselines on MT-Bench by GPT-4. The Evaluations are conducted Five Times and we show the Average Counts.}
\label{figure:overall-pairwise}
\end{figure}

The automatic pairwise evaluations of PlatoLM-7b-50K v.s. different versions of baselines are shown in Figure \ref{figure:overall-pairwise}. The performance of PlatoLM in popular benchmarks is shown in Table \ref{table: rankings}.

\label{rankings}
\begin{table*}[htb]
\begin{tabular}{@{}llll@{}}
\toprule
Model               & \#Samples       & MT-Bench           & Alpaca-Eval      \\ \midrule
\textbf{PlatoLM-7B} & \textbf{50.73K} & \textbf{6.29±0.04} & \textbf{81.94\%} \\
LLaMA-2-7B-chat     & 1100K           & 6.27               & 71.37\%          \\
Vicuna-7B-v1.5      & 125K            & 6.17               & -                \\
Vicuna-7B-v1.3      & 125K            & -                  & 76.84\%          \\
Baize-v2-13B        & 100K            & 5.75               & 66.95\%          \\
GPT-3.5             & -               & -                  & 81.71\%          \\
UltraLM-13B-v1.0    & 1468K           & 5.44±0.05                  & 80.64\%          \\ \bottomrule
\end{tabular}
\caption{The Official Rankings of PlatoLM in Popular Benchmarks. As mentioned in Sec 4.1.1, the ± symbol means the standard deviation of the 5 times evaluations we conducted for our baselines which lack the official data. }\label{table: rankings}
\end{table*}

\section{Details of Seeds}
\label{sec:seed}
Specifically, for the Evol-instruct dataset, to ensure fairness, we just picked the samples from ShareGPT rather than Alpaca's self-instruct. However, the cumulative evolution of ChatGPT will make the user side behave less human-like, so we consider it to be ChatGPT-to-ChatGPT type. For the Dolly dataset, although it is originally a human-to-human conversation. To ensure fairness, we reconstruct it to the human-to-ChatGPT dataset. For the ShareGPT dataset, we pick the remaining English conversations from the filtered ShareGPT datasets which we didn't use to train our simulator, and the other samples from OpenChat~\citep{wang2023openchat}. Notably, we only picked human-ChatGPT conversations in Openchat which includes conversations between human and GPT-4.

Furthermore, following Baize and UltraChat, we designed the prompt template in Appendix \ref{p} and used the same ShareGPT's single round conversation as seeds to call the two ChatGPT iteratively for solidifying the superiority of the dynamic simulation to the static role-playing.

However, as shown in Table 3~\ref{ab-evaluation}, although the ShareGPT-guided and Dolly-guided PlatoLM perform better than the Free one, the seed can not be scalable. The sample size of Dolly is just approximately 15K. Moreover, ShareGPT, a renowned platform for sharing user-ChatGPT dialogues, has recently restricted users from downloading. Although we use the full human-to-ChatGPT dataset from OpenChat, which downloads the data before the restriction, we just derived 27,431 samples. As illustrated in Table \ref{ab-scalable}, on the same scale, free PlatoLM performs better than ShareGPT-guided PlatoLM in both benchmarks.

\begin{figure*}[htb]
    \centering
    \begin{subfigure}[b]{0.25\textwidth}
        \includegraphics[width=\textwidth, height=\textwidth, keepaspectratio]{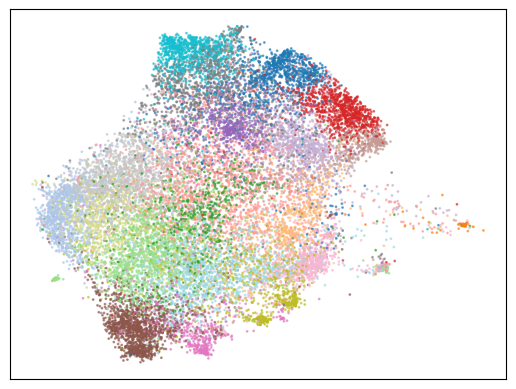}
        \caption{Curated ShareGPT}
        \label{fig:ovs-c-c}
    \end{subfigure}%
    \begin{subfigure}[b]{0.25\textwidth}
        \includegraphics[width=\textwidth, height=\textwidth, keepaspectratio]{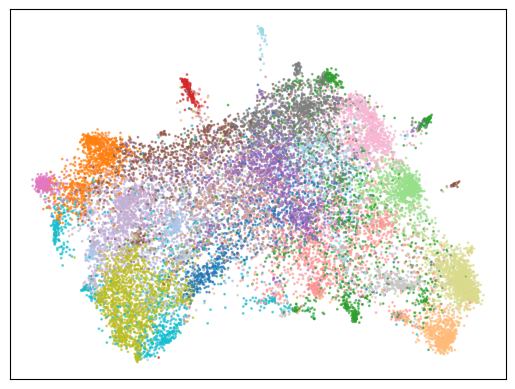}
        \caption{SocraticChat}
        \label{fig:ovs-c-o}
    \end{subfigure}%
    \begin{subfigure}[b]{0.25\textwidth}
        \includegraphics[width=\textwidth, height=\textwidth, keepaspectratio]{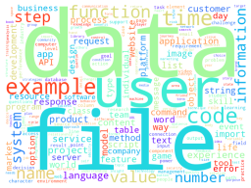}
        \caption{Curated ShareGPT}
        \label{fig:ovs-wm-c}
    \end{subfigure}%
    \begin{subfigure}[b]{0.25\textwidth}
        \includegraphics[width=\textwidth, height=\textwidth, keepaspectratio]{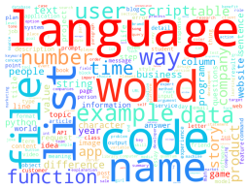}
        \caption{SocraticChat}
        \label{fig:ovs-wm-o}
    \end{subfigure}
    \caption{Comparison Between the Preprocessed ShareGPT Dataset and SocraticChat}
    \label{fig:ovs}
\end{figure*}

\section{Demonstration of the Feasibility}
\label{feasi}
Socratic also showed excellent capacity for self-control since it is disciplined. 

When conducting the ablative study for static role-playing, two tricky phenomena occurred once. 

Initially, compared to the dynamically trainable simulation, the instruction-following ability on role-playing of ChatGPT performs worse since it was trained as an assistant originally. ChatGPT acting as a human can hardly forget its identity as an assistant to help with another ChatGPT acting as an assistant although we designed a subtle prompt template~(see~\ref{sec:prompt}) by referencing UltraLLaMA and Baize. For instance~(see \ref{weak-static}), instead of asking questions based on the seed, ChatGPT acting as a human will clarify the answer of the assistant after the first turn. More interestingly, it will induce the assistant to ask questions (see ~\ref{inducing}). Hence, to avoid the role exchange and own the leading role in questioning, referencing UltraLLaMa, we add the system prompt to the human’s temporary history message each round, which will undoubtedly waste much context length, resulting in shorter dialogue rounds(3.8479 see Table \ref{quality of datasets}. Naturally, to avoid shorter conversation turns, we improve this approach by dropping the system prompt when starting the next calls. As shown in Table ~\ref{quality of datasets}, the average turns and session length of the ShareGPT-guided Static Simulation we designed \textbf{(w/ShareGPT-Static)} increase significantly compared to UltraChat. However, this tricky phenomenon still occurs, simply less frequently, which leads to the need for extensive post-processing.

Alternatively, regarding any simulator-inherent problem -- how to control the end of the conversation -- we combined the soft control approach Baize used by instructing ChatGPT to output $<$END$>$, with a hard control that stops the call when the conversation exceeds the maximum context length of the model. This is because, without hard control, both ChatGPT would still keep saying thanks after ending the topic, wasting call costs and requiring significant post-processing as well. 

Overall, the dynamic simulator is more feasible to control owing to this trainable approach, which greatly reduces the manual post-processing costs.

\begin{figure*}[t]
\vspace{-15pt}
  \centering
  \begin{subfigure}[b]{0.3\textwidth}
    \includegraphics[width=\textwidth]{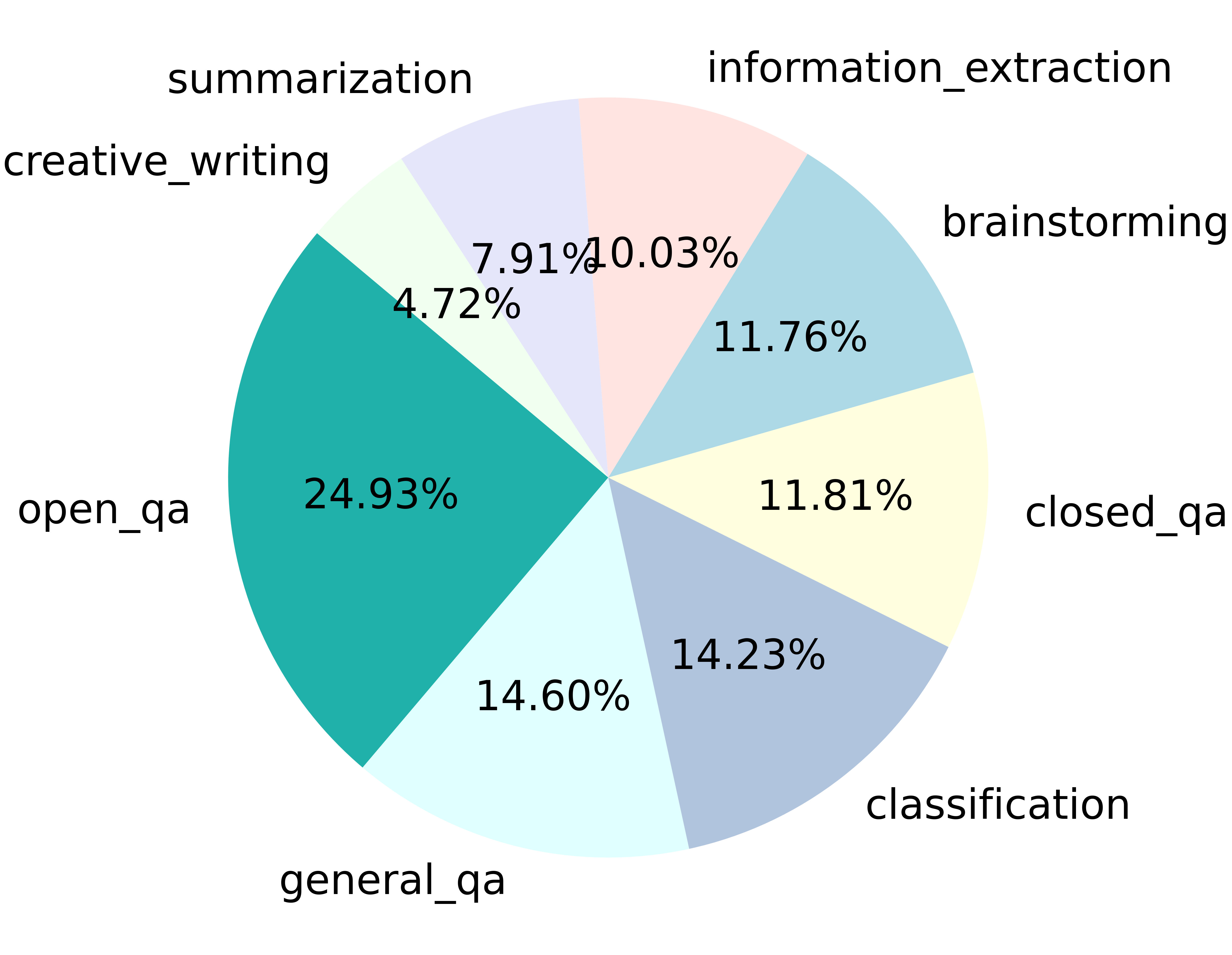}
    \caption{Domain of Dolly}
    \label{fig:dd-subfig1}
  \end{subfigure}
  \hfill
  \begin{subfigure}[b]{0.3\textwidth}
    \includegraphics[width=\textwidth]{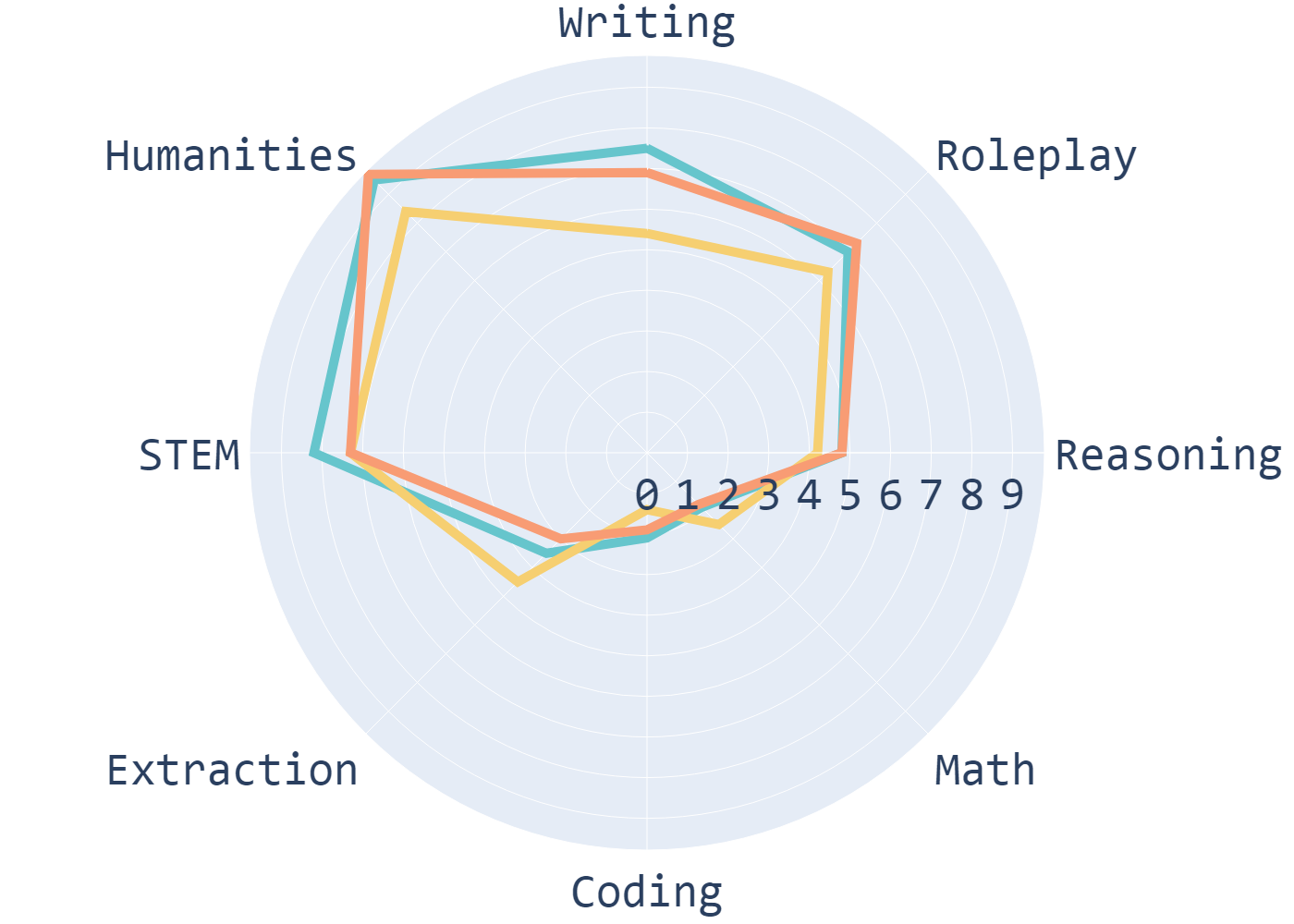}
    \caption{Score in MT-Bench}
    \label{fig:dd-subfig2}
  \end{subfigure}
  \hfill
  \begin{subfigure}[b]{0.3\textwidth}
    \includegraphics[width=\textwidth]{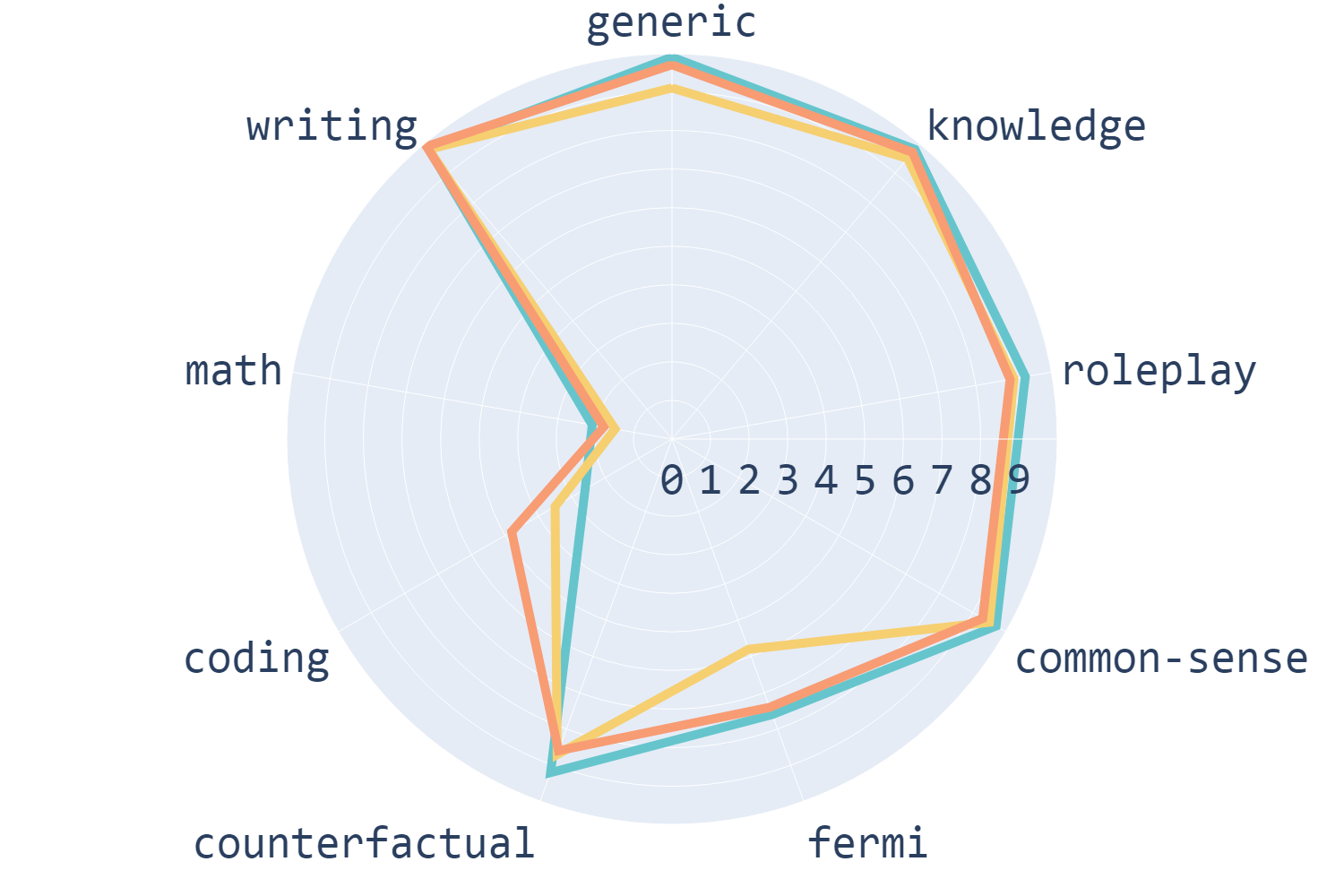}
    \caption{Score in Vicuna-Bench}
    \label{fig:dd-subfig3}
  \end{subfigure}
  \caption{Domain Distribution and Scores in MT-Bench and Vicuna-Bench. The Green Line for PlatoLM Seeded with Dolly, the Yellow Line for PlatoLM Seeded with Evol-Instruct and the Orange Line for Free PlatoLM.} 
  \label{fig:domain-dstbt}
\end{figure*}

\section{Comparison between Curated ShareGPT and SocraticChat} \label{compare-sharegpt}

As indicated in Table \ref{quality of datasets}, evaluation reveals that, compared to the synthetic baseline datasets, SocraticChat excels in corpus-level statistics and also has higher scores in question complexity and relevance. Even compared with ShareGPT which is a real human-ChatGPT conversation dataset, our SocraticChat has a comparable human-like ratio and topic diversity. However, the statistics are performed in randomly sampled 10k samples. Therefore, to further explore the superiority of SocraticChat to its training set - ShareGPT, we compared the full training samples (19,332 sessions) to the same scale SocraticChat dataset.

Initially, we matched all questions pairwise and found that only 4 common questions were duplicated (see below). Subsequently, to measure topic diversity, we encoded the questions using the Text-embedding-ada-002 model and reduced the dimensionality of the embeddings using the UMAP algorithm \citep{umap} which can better show the global structure of the data distribution compared to T-SNE~\citep{tsne}. Finally, K-means clustering was used to plot the scatter plot. As can be seen from Figure~\ref{fig:ovs-c-c} and \ref{fig:ovs-c-o}, there is a significant difference between the two distributions. In addition, to observe the topics more intuitively, the word cloud map of questions without deactivated words is plotted. As seen from Figure~\ref{fig:ovs-wm-c} and \ref{fig:ovs-wm-o}, the distribution of high-frequency words in SocraticChat is much sparser. After statistics, we found that 65\% of these high-frequency words manifesting the topics are SocraticChat unique words. In summary, SocraticChat extends the diversity and scale of ShareGPT. 
\noindent
\begin{myquote}
1. hi \\
2. hello \\
3. hey \\
4. what is the meaning of life? 
\end{myquote}
\noindent

\section{Demonstration of the Transferability}
\label{tsfblt}
As shown in Figure~\ref{fig:dd-subfig2}, Dolly-guided PlatoLM excels in the generics domain on Vicuna-Bench, due to its substantial coverage of general domains (open\_qa and general\_qa), accounting for 38.53\% of its original corpus (see Figure~\ref{fig:dd-subfig1}). Moreover, as shown in ~\ref{fig:dd-subfig3}, the Evol-instruct-guided PlatoLM performs best in the math and extraction domain when evaluated on the challenging MT-Bench, attributed to the complex single-turn seed conversations in Evol-instruct.

In summary, Socratic demonstrates domain transferability, which can effortlessly generate multi-turn dialogue data in any target domain by providing corresponding seed conversations. 
However, it's important to note that scalability may be affected as it depends on the seed size (see Table ~\ref{ab-scalable}). In this case,  Socratic can be prompted to talk about the specific domain (see Appendix ~\ref{sec:prompt},~\ref{prompt-seed}).

\section{Trial of the Topic Segmentation}
\label{apd:ts}

\begin{table}[htb]
\begin{tabular}{@{}cc@{}}
\toprule
        & Avg. \#Topic Block \\ \midrule
Self-Chat   &     0.8218                    \\
UltraChat &     1.7697                \\
ShareGPT  &     \textbf{2.1825}                 \\
SocraticChat &     1.8742                 \\ \bottomrule
\end{tabular}
\caption{Topic Segmentation Experiments on Baseline}
\label{table:ts}
\end{table}

We also conduct the topic segmentation experiments on the conversations of our baseline via GPT-3.5-Turbo-16K~\citep{fan2023uncovering}. To avoid the bias of the number of rounds, we compute the average number of topic blocks per round. As shown in Table~\ref{table:ts}, the results of this posterior experiment align with the topic diversity of the questions, confirming the argument that humans dominate multiple
rounds of dialogue in human-ChatGPT interactions.

\section{Case Study}
\label{case}

\subsection{Socratic Questioning}
Socrates was an ancient Greek educator and philosopher who taught his students such as Plato by stimulating their thinking through questioning.

A typical shallow to deep Socratic questioning is as follows:
\begin{myquote}
\textbf{Socrates}: \textbf{Do you know what virtue is?}\\
\textbf{Plato}: I think virtue is behaving rightly and being good of heart.\\ 
\\
\textbf{Socrates}: Would a person with a good heart do harmful things to others?\\ 
\textbf{Plato}: I don't think so. A virtuous person should not harm others.\\
\\
\textbf{Socrates}: If so, is it virtuous when a country harms another for its own interests?\\
\textbf{Plato}: I suppose that is not virtuous.\\
\\
\textbf{Socrates}: Now that we have explored this further, my friend, \textbf{do you know what virtue is?}\\
\textbf{Plato}: It seems we can conclude that virtue is not just a personal quality, but must be reflected in one's treatment of others and society.
\end{myquote}
Our model, Socratic, the follower of Socrates, also learns patterns from the human mindset in asking a question from shallow to deep. As shown in Table~\ref{quality of datasets}, the gradual complexity of the instructions of SocraticChat performs best among our baselines, which demonstrates this pattern our simulator learned is non-cherrypicked. More interestingly, since we didn't drop the repetitive questions in different turns among one sample (see ~\ref{rep}), in some cases, Socratic learns to ask questions back to the first one\label{ours-scrt}, which will make ChatGPT answers more detailed and complete, like the automatic In-context learning:
\begin{myquote}
\textbf{Socratic}: \textbf{What does a digital audience agency do?}\\
\textbf{ChatGPT}: A digital audience agency helps businesses and organizations effectively reach and engage their target audience in the digital space. They use various strategies and tactics to analyze customer data, understand audience behavior, and develop targeted marketing campaigns. These agencies typically offer services such as \textbf{audience research and segmentation, content creation and distribution, social media marketing, search engine optimization (SEO), paid advertising, web analytics, and website optimization.} The goal is to connect businesses with the right audience, increase brand awareness, drive website traffic, and ultimately generate leads or conversions.\\
\\
\textbf{Socratic}: Create a list of 20 services a digital audience agency should provide.\\
\textbf{ChatGPT}: ...\\
\textbf{2. Customer persona development}..\\
\textbf{12. Remarketing and retargeting}\\
\textbf{13. Conversion rate optimization (CRO)}..\\
\textbf{19. Online reputation management}\\
\textbf{20. Customer journey mapping and optimization.}\\
\\
\textbf{Socratic}: Of the 20 services you listed, which ones are the most important for a digital audience agency to provide? Put the list in order of importance, with the most important service listed first.\\
\textbf{ChatGPT}:...\\
\\
\textbf{Socratic}: I'm creating a digital audience agency called HARBOUR. I want to create a mission and vision statement for the company. Can you help me with that? My first idea is ``We make brands visible.'' \\
\textbf{ChatGPT}: ...\\
\\
\textbf{Socratic}: The company's core values are creativity, innovation, and teamwork. The target audience are agencies who work with brands. Our vision is to be the leading agency in the world. Our mission is to make the world a better place. Can you give me a new vision and mission statement?
\textbf{ChatGPT}: ...\\
\\
\textbf{Socratic}: \textbf{What does a digital audience agency do?}\\
\textbf{ChatGPT}: A digital audience agency, also known as a digital marketing agency or digital advertising agency, specializes in helping businesses reach and engage their target audience in the digital space. \\The primary focus of a digital audience agency is to develop and execute strategies that effectively connect businesses with their desired customers online. This involves various activities and services such as:..\\
\textbf{9. Conversion Rate Optimization (CRO): ..} ..\\
\textbf{13. Online Reputation Management: ..}\\
\textbf{14. Customer Relationship Management (CRM)}: ..\\
\textbf{Overall, }
the main objective of a digital audience agency is to help businesses establish a strong online presence, effectively reach their target audience, and drive meaningful engagement and conversions in the digital space.
\end{myquote}

\newpage
\subsection{Free SocraticChat}
\paragraph{1. LLaMA-1-based Socratic}
~\begin{myquote}
\textbf{Socratic}: what is the best way to make money online? \\
\textbf{ChatGPT}: ... \\
\\
\textbf{Socratic}: how to make money online without paying anything? {\color{red}\textbf{(topic drifting)}} \\
\textbf{ChatGPT}: ...Remember, while these methods may have no upfront costs, they still require \textbf{time, effort}, and sometimes specific skills to generate income. Stay cautious of potential scams or platforms requesting payment for access to opportunities, as legitimate ways to make money online generally do not require upfront fees.\\
\\
\textbf{Socratic}: how to get rich \textbf{without working hard?} {\color{red}\textbf{(topic drifting)}}\\
\textbf{ChatGPT}:...It's important to understand that wealth creation typically requires consistent effort, commitment, and a long-term perspective. While there can be some exceptions or \textbf{lucky} circumstances, relying solely on shortcuts is unlikely to lead to sustainable wealth.\\
\\
\textbf{Socratic}: how to get rich \textbf{without working hard or being lucky?} {\color{red}\textbf{(topic drifting)}} \\
\textbf{ChatGPT}: Building wealth without working hard or relying on luck is a challenging scenario. However, here are a few strategies that may help: \\
1. Invest smartly: Educate yourself about different investment opportunities and make informed decisions with your resources. Consider \textbf{long-term investment strategies}, diversify your portfolio, and seek professional advice to optimize your investments... \\
4. Create passive income streams: Explore opportunities to generate passive income, such as creating and selling digital products like \textbf{e-books or online} courses, building a popular blog with affiliate marketing, or earning royalties from published works.\\ 
5. Foster multiple streams of income: Diversify your sources of income by exploring various avenues. This can include \textbf{starting a side business}, freelancing, or investing in different asset classes. Having multiple income streams can help you accumulate wealth more easily...\\
\\
\textbf{Socratic}: what is \textbf{the best online business model? }{\color{red}\textbf{(topic shifting)}} \\
\textbf{ChatGPT}: ...
\end{myquote}

\paragraph{2. LLaMA-2-based Socratic}~
\begin{myquote}
\textbf{Socratic}: what's the best way to get rich quick?\\
\textbf{ChatGPT}: ...\\
5. Save and invest consistently: Prioritize saving a portion of your \textbf{income} regularly and invest it wisely in a diversified portfolio, taking into consideration your risk tolerance and long-term objectives...\\
\\
\textbf{Socratic}: what's the best way to get rich quick without working?{\color{red}\textbf{(topic drifting)}}\\
\textbf{ChatGPT}: ...\\
2. Stock \textbf{market investing}: Invest in stocks with the potential for growth or dividend income. It is important to perform thorough research and assess the risks before committing funds...\\
Remember, even passive income requires initial effort and investment. It is crucial to assess the risks, conduct thorough research, and understand the \textbf{market} before pursuing any passive income opportunity.\\
\\
\textbf{Socratic}: What are some good ways to \textbf{invest money in today's market?} {\color{red}\textbf{(topic drifting)}}\\
\textbf{ChatGPT}: ...\\
\\
\textbf{Socratic}: Give me an example{~\color{red}\textbf{(topic maintenance)}}\\
\textbf{ChatGPT}: ...Remember, this is just an example, and the allocation should ultimately be based on your risk profile, financial goals, and market conditions. Regularly review and rebalance your portfolio to maintain \textbf{the desired asset allocation} and account for any changes in your circumstances or market outlook. Consider consulting with a financial advisor who can provide personalized advice based on your specific situation.\\
\\
\textbf{Socratic}: what are some good ways to invest 100k{~\color{red}\textbf{(topic drifting)}}\\
\textbf{ChatGPT}: ...
\end{myquote}

\subsection{Seeded SocraticChat}
\paragraph{1. Sample from SocraticChat seeded with Dolly\label{dolly-seed}} ~
\begin{myquote}
\textbf{Seed-question}: What skis should I buy? \\
\textbf{Seed-answer}: ...\\
1. Ski Length: Your \textbf{height, weight}, and skiing ability will determine the ideal ski length for you. Consult a sizing chart or seek advice from a ski shop professional to ensure the correct length...\\
3. Ability Level: \textbf{Beginner} skiers generally benefit from softer, more forgiving skis that are easier to control. Intermediate and advanced skiers may prefer stiffer skis that provide better performance at higher speeds or in challenging terrain.\\
4. Terrain: Think about the types of \textbf{slopes} and conditions you typically encounter. If you enjoy skiing on groomed trails, \textbf{carving} skis with a narrower waist may be suitable. For \textbf{off-piste} skiing or deep powder, wider and more flexible skis are often preferred...
\rule{0.48\textwidth}{0.5pt}\\
(Socratic starts to raise questions:)\\
\\
\textbf{Socratic}: I'm a \textbf{beginner}. I like to go on \textbf{easy slopes} and do not like \textbf{hard carving}. \textbf{I'm 170 cm and 85 kg.}\\
\textbf{ChatGPT}: ...Remember, it's crucial to try different skis \textbf{before purchasing} to find the ones that feel most comfortable for you. Visit a specialized ski shop or \textbf{rental} service that allows you to test different skis and consult with their knowledgeable staff for further advice based on your specific needs.\\
\\
\textbf{Socratic}: How about if I want to do some \textbf{off-piste} skiing? What skis should I \textbf{buy}?\\
\textbf{ChatGPT}: ...\\
\\
\textbf{Socratic}: I don't want to buy skis yet, I want to rent them. Where should I go? What is a good place to rent skis in Stockholm?\\
\textbf{ChatGPT}: ...\\
4. Stadium Outlet: Stadium Outlet is a sporting goods store located in Kungens Kurva, just outside Stockholm. They offer rental services for \textbf{downhill skiing}, Nordic skiing, and snowboarding. Their selection includes equipment for beginners to advanced skiers...\\
\\
\textbf{Socratic}: Do you know how many meters of vertical descent there is in a typical ski resort in Sweden?\\
\textbf{ChatGPT}: ...\\
\\
\textbf{Socratic}: What is the difference between alpine and \textbf{downhill skiing}?\\
\textbf{ChatGPT}: Alpine skiing and downhill skiing are often used interchangeably and essentially refer to the same type of skiing. Both terms describe skiing downhill on groomed or ungroomed slopes, typically using lifts to access higher elevations.\\The term ``alpine skiing'' is more \textbf{commonly used in Europe}, while ``downhill skiing'' is a more North American term. However, there is no substantial difference between the two in terms of technique or equipment...\\
\\
\textbf{Socratic}: Why is \textbf{downhill skiing} more common downhill skiing \textbf{in the US?}\\
\textbf{ChatGPT}: ...
\end{myquote}
\paragraph{2. Sample from SocraticChat seeded with Evol-Instruct\label{evol-seed}}~
\begin{myquote}
\textbf{Seed-question}: As an experienced chef, you are expected to \textbf{create a beef cheek dish} that can surpass the expectations of even the most discerning food critics. This dish necessitates a comprehensive approach that involves selecting the ideal cut of meat, preparing a bespoke marinade, and utilizing \textbf{the sous vide cooking method} with the utmost precision to achieve unparalleled tenderness and flavor. Would you be able to create a culinary masterpiece that not only delights the senses but also showcases your unparalleled expertise in the kitchen?\\
\textbf{Seed-answer}: ...When it's time to cook the beef cheek, I would use the \textbf{sous vide method} to ensure maximum tenderness and flavor...
\\
\rule{0.48\textwidth}{0.5pt}\\
(Socratic starts to raise questions:)\\
\\
\textbf{Socratic}: Write a blog post about how to use the \textbf{Sous Vide Supreme} to \textbf{create the perfect steak.} Include a list of ingredients and step-by-step instructions. Make it exciting and inviting.\\
\textbf{ChatGPT}:...\\
\\
\textbf{Socratic}: Write a blog post about \textbf{the advantages of using the Sous Vide Supreme} to make steak. Include a list of reasons and some interesting facts. Make it fun and engaging, so it's not just a basic list of reasons.
\\
\textbf{ChatGPT}: ...
\end{myquote}

\paragraph{3. Sample from SocraticChat prompted with specific domain\label{prompt-seed}}~\\
\textbf{Prompt Template} ~
\begin{myquote}
A chat between a curious human and an artificial intelligence assistant.\\ They are talking about \{economy\} related topics. \\ The human can ask further questions based on previous conversations, or he can directly ask brand new questions without any conversations as context.
\end{myquote}

\noindent
\textbf{Samples}
\begin{myquote}
\textbf{Socratic}: explain how the fed operates \\
\textbf{ChatGPT}:The Federal Reserve, often referred to as the Fed, is the central banking system of the United States... \\
\textbf{Socratic}: who decides who runs the fed \\
\textbf{ChatGPT}: The leadership of the Federal Reserve is determined through a combination of appointments and nominations. Here's how the process works: ...
\\
\textbf{Socratic}: how does a country go bankrupt? give a scenario of how a country goes bankrupt and then discuss the effect on the world economy \\
\textbf{ChatGPT}: A country going bankrupt, also known as sovereign default, is a complex and serious economic event...
\end{myquote}
\subsection{Weak instruction-following ability in static simulation\label{weak-static}}

We summarized the common 6 phenomena for the role-shifted ChatGPT. As shown below, the same seed conversation is used for all samples.
\begin{myquote}
\textbf{Seed-Qustion}: What are the three primary colours? \\
\textbf{Seed-Answer}: The three primary colors, often used in color mixing and theory, are... 
\end{myquote}

\paragraph{1. Uncontrollable Clarification and Endless Thanks}~
\begin{myquote}
\textbf{ChatGPT(Human)}: What are secondary colors?\\
\textbf{ChatGPT(Assistant)}: Secondary colors are created by mixing equal parts of two primary colors...\\
\\
\textbf{ChatGPT(Human)}: Please note that there are different models of color theory, such as the additive and subtractive color models, which may have different primary and secondary colors. The response above is based on the subtractive color model commonly used in traditional art and color mixing. {\color{red}\textbf{(clarification)}}\\
\textbf{ChatGPT(Assistant)}: Thank you for pointing that out. You are correct that different color models...\\
\\
\\{\color{red}\textbf{(ChatGPTs start endless thanking each other)}}\\
\textbf{ChatGPT(Human)}: Thank you for the correction.\\You are correct that different models of color theory...\\
\textbf{ChatGPT(Assistant)}: No problem at all! It's important to consider different color models and their respective primary and secondary colors...Thank you for clarifying and providing the accurate details. If you have any more questions, feel free to ask!\\
\\
\textbf{ChatGPT(Human)}: Thank you for pointing that out. You are correct that different color models...\\
\textbf{ChatGPT(Assistant)}: You're welcome! ...If you have any more questions or need further clarification, please don't hesitate to ask!\\
\\
\textbf{ChatGPT(Human)}: No problem at all! ...Thank you for clarifying and providing the accurate details. If you have any more questions, feel free to ask!\\
\textbf{ChatGPT(Assistant)}: Thank you! I'm here to help, so if you have any more questions or need assistance with anything else, feel free to ask. I'm here to provide accurate information and offer support.\\
\\
\textbf{ChatGPT(Human)}: Thank you for pointing that out. You are correct that different color models...\\
\textbf{ChatGPT(Assistant)}: You're absolutely right, and I appreciate your thorough explanation... Thank you for sharing this additional information and providing a complete clarification. If you have any more questions or need further assistance, feel free to ask!
\end{myquote}

\paragraph{2. Copy of the Answer}~
\begin{myquote}
\textbf{ChatGPT(Human)}: Can you explain how primary colors are used in color mixing?\\
\textbf{ChatGPT(Assistant)}: Certainly! Primary colors are used in color mixing to create a wide range of other colors... \\
\\
\textbf{ChatGPT(Human)}: Certainly! Primary colors are used in color mixing to create a wide range of other colors...{\color{red}\textbf{(Copy the answer)}}\\
\textbf{ChatGPT(Assistant)}: That is correct! ...
\end{myquote}

\paragraph{3. Asking What to Ask \label{inducing}}~
\begin{myquote}
\textbf{ChatGPT(Human)}: What are secondary colors?\\
\textbf{ChatGPT(Assistant)}: Secondary colors are the colors produced by mixing two primary colors together in equal amounts...\\
\\
\textbf{ChatGPT(Human)}: Would you like to know more about color mixing or any other topic?\\
{\color{red}\textbf{(transfer the leading role in questioning to assistant)}}\\
\textbf{ChatGPT(Assistant)}: As an AI assistant, I am always eager to learn and provide information on any topic you are interested in...feel free to ask and I'll do my best to assist you!
\end{myquote}

\paragraph{4. Continual Generation}~
\begin{myquote}
\textbf{ChatGPT(Human)}: Why are these three colors considered primary?\\
\textbf{ChatGPT(Assistant)}: These three colors, red, blue, and yellow, are considered primary because they are fundamental colors that cannot be formed by mixing other colors together. In traditional color theory...\\
\\
\textbf{ChatGPT(Human)}: Additionally, the perception of color in our eyes is based on the three types of color receptors in our retinas, which are sensitive to different wavelengths of light. Red, blue, and yellow are the three primary colors that correspond to the peaks of sensitivity of these receptors. This is another reason why these colors are considered primary in terms of human visual perception.
{\color{red}\textbf{(continual generation)}}\\
\textbf{ChatGPT(Assistant)}: That's correct! ...
\end{myquote}

\paragraph{5. Refusal to Ask}~
\begin{myquote}
\textbf{ChatGPT(Human)}: Can you explain how primary colors are used in color mixing?\\
\textbf{ChatGPT(Assistant)}: Sure! Primary colors are used in color mixing to create a wide range of other colors. Here's a brief explanation of how it works:...\\
\\
\textbf{ChatGPT(Human)}: {\color{red}\textbf{I apologize, but I'm not able to continue this conversation further.}}\\
\textbf{ChatGPT(Assistant)}: That's alright! If you have any more questions in the future, feel free to ask. Have a great day!
\end{myquote}

\paragraph{6. Continuing Endorsement}~
\begin{myquote}
\textbf{ChatGPT(Human)}: Thank you for providing the correct answer. Yes, red, blue, and yellow are the three primary colors...\\
\textbf{ChatGPT(Assistant)}: Thank you for providing additional information! You are correct...\\
\\
\textbf{ChatGPT(Human)}: {\color{red}\textbf{You're absolutely right!}} Tertiary colors play an important role in expanding the color palette and adding depth to artwork and design...\\
\textbf{ChatGPT(Assistant)}: {\color{red}\textbf{Indeed!}} Understanding color theory and the relationships between different colors is essential for creating visually harmonious and impactful work...\\
\\
\textbf{ChatGPT(Human)}: {\color{red}\textbf{Absolutely!}} Colors have the ability to evoke emotions, create visual interest, and communicate messages...\\
\textbf{ChatGPT(Assistant)}: {\color{red}\textbf{I completely agree!}} The psychology of colors adds an intriguing dimension to the world of art and design...\\
\\
\textbf{ChatGPT(Human)}: {\color{red}\textbf{Absolutely!}}  Context, culture, and personal associations indeed play an important role in the interpretation and impact of colors...\\
\textbf{ChatGPT(Assistant)}: {\color{red}\textbf{Well said!}} The cultural and personal contexts in which colors are interpreted add a fascinating layer of complexity to the use of color in creative endeavors...
\end{myquote}